\documentclass[runningheads]{llncs}
\usepackage{subfigure}
\usepackage{booktabs}
\usepackage{bbding}
\usepackage{wrapfig}
\usepackage{adjustbox}
\usepackage{graphicx}
\usepackage{algorithm} 
\usepackage{algpseudocode}
\usepackage{tabularx}
\usepackage{xcolor,colortbl}
\usepackage{verbatim}
\usepackage{float}
\usepackage{makecell}
\usepackage{cite}
\usepackage{sidecap}
\usepackage{wrapfig}

\usepackage{hyperref}

\definecolor{Gray}{gray}{0.85}
\definecolor{LightCyan}{rgb}{0.88,1,1}

\usepackage{array}
\newcolumntype{a}{>{\columncolor{Gray}}m{2cm}}
\newcolumntype{b}{>{\columncolor{white}}m{2cm}}
\newcolumntype{d}{>{\columncolor{white}}m{2.5cm}}
\newcolumntype{e}{>{\columncolor{Gray}}m{2.1cm}}
\newcolumntype{f}{>{\columncolor{white}}m{2.1cm}}

\begin{document}
\title{Domain Generalizer: A Few-shot Meta Learning Framework for Domain Generalization in Medical Imaging}
\titlerunning{Modality Generalizer: Domain Generalization in Medical Imaging}
%

\author{Pulkit Khandelwal\inst{1,2} \Envelope \and
Paul Yushkevich\inst{2}}

%
\authorrunning{P. Khandelwal and P. Yushkevich}

\institute{Department of Bioengineering, University of Pennsylvania, Philadelphia, PA, United States \and
Penn Image Computing and Science Laboratory, Department of Radiology, University of Pennsylvania, Philadelphia, PA, United States\\
\email{pulks@seas.upenn.edu}}
\maketitle           

\begin{abstract}
Deep learning models perform best when tested on target (test) data domains whose distribution is similar to the set of source (train) domains. However, model generalization can be hindered when there is significant difference in the underlying statistics between the target and source domains. In this work, we adapt a domain generalization method based on a model-agnostic meta-learning framework \cite{MLDG} to biomedical imaging. The method learns a domain-agnostic feature representation to improve generalization of models to the unseen test distribution. The method can be used for any imaging task, as it does not depend on the underlying model architecture. We validate the approach through a computed tomography (CT) vertebrae segmentation task across healthy and pathological cases on three datasets. Next, we employ few-shot learning, i.e. training the generalized model using very few examples from the unseen domain, to quickly adapt the model to new unseen data distribution. Our results suggest that the method could help generalize models across different medical centers, image acquisition protocols, anatomies, different regions in a given scan, healthy and diseased populations across varied imaging modalities.

\end{abstract}

\section{Introduction and Background}
\label{Section1}
In biomedical imaging, deep learning models trained on one dataset are often hard to generalize to other related datasets. Generally, biomedical images can be represented as points on a high-dimensional non-linear manifold. Failure of segmentation and classification algorithms to generalize across imaging modalities, patients, image acquisition protocols, medical centers, healthy and diseased populations, age, etc., can be explained by significant differences in the statistical distributions of datasets on the image manifolds, known as \emph{covariate shift} \cite{covariate}. Addressing covariate shift by retraining deep learning models on each new data domain is impractical in most applications because of the scarcity of expert labeled data. Therefore, it is important to develop deep learning methods that generalize well to new related datasets not seen during training using few or no annotated examples from the new dataset. \textit{Domain adaptation} \cite{DANN} and \textit{domain generalization} \cite{DG_pacs} paradigms aim at reducing the covariate shift between the training and test distributions by learning domain invariant features. Domain adaptation learns a feature representation that is invariant to the statistics of the source and target domains, and is discriminative enough for the actual learning task. Domain adaptation could either be unsupervised or semi-supervised. Domain generalization, a relatively less studied and harder problem, trains models using a variety of source domains to learn a generic feature representation which should perform well on unseen target domains. This flavor of transfer learning does not use any samples from the target distribution during training. Relatedly, few-shot learning is a paradigm which adapts a trained model to a completely new data distribution with very limited labeled training examples \cite{fewshot}.

\textit{The biomedical imaging} community has witnessed several applications of domain adaptation and few-shot learning. Adaptation across different medical centers is a known challenge in image segmentation \cite{glocker_site}, and has been achieved through both unsupervised \cite{unsupervised}, and supervised approaches \cite{supervised}. Cross-modality domain adaptation methods between magnetic resonance (MR) and CT images have been proposed using variational autoencoders \cite{crossmodality} for whole heart segmentation, and CycleGANs for segmentation of the prostate \cite{cycleGAN}, \cite{pnp}. Decision forests have been employed to adapt between in-vivo and in-vitro images \cite{invitro} for intravascular ultrasound tissue segmentation. A few-shot network \cite{fewshotSqueeze} was proposed to segment multiple organs in MR images. However, a priori knowledge of the unseen test domain is not always available, which hinders model generalizability as discussed above.

Very recently, some groups explored domain generalization for biomedical imaging. (1) A series of nine data augmentation techniques were applied to the training domains to mimic the test distribution \cite{stacked} for heart ultrasound, heart and prostate MR image segmentation. (2) An episodic training-based meta-learning method \cite{glocker_gen} was applied to segment brain tissue in T1-weighted MRI across four medical centers. (3) A variational auto-encoder \cite{diva} was used to learn three latent subspaces to generalize across patients for a 2D cell segmentation task via domain disentanglement. These methods have certain limitations respectively: (1) In \cite{stacked}, the training data is very large and heavily augmented, which might not be the general case for many problems in medical imaging where the goal is to extract enough information from very limited data for domain generalization; the method is not tested on diseased populations which could vary significantly in anatomies and shapes from a generic healthy population. (2) In \cite{glocker_gen}, again, the training set contains similar anatomies in both the train and test sets; the average performance is quite marginal than the compared baseline, with an improvement of only around 0.8\% in Dice score (Baseline: 90.6\% vs proposed: 91.4\%); does not evaluate on cases with atrophied or irregular brain anatomies. (3) In \cite{diva}, the method uses domain labels as an additional cue, which we argue is difficult to define precisely, due to its wide range of interpretation; the average performance is not significant with an improvement of 0.4\% in average Dice score (Baseline: 95.4\% vs proposed: 95.8\%); the method evaluates only one 2D dataset, not extending to the 3D case; each patient is considered as  different domain acquired at the same medical center, and hence the test and train sets might have similar statistical distributions.

\textbf{Contributions.} In the present work, we extend a gradient-based meta-learning domain generalization method (MLDG) \cite{MLDG} that has shown promise for image classification tasks to the context of biomedical image segmentation, termed \textbf{MLDG-Seg}. To evaluate this approach, we focus on the problem of vertebrae segmentation in CT images and utilize three publicly available databases. To address the above-mentioned drawbacks of existing domain generalization methods, we construct three domain generalization contexts: (a) \textit{generalization to new anatomies}: the vertebrae are divided into four domains: lumbar, lower, middle, and upper thoracic regions. The model is trained on three domains, and then tested on the unseen fourth domain. (b) \textit{generalization to a diseased population with fractured vertebrae, dislocated discs}: the model is trained on a healthy population, then tested on unseen data of a diseased population from \textit{another medical center}. (c) \textit{generalization to unseen anatomies, surgical implants, different acquisition protocols, arbitrary orientations, and field of view (FoV)}: the model is tested on a very large dataset. Through these three contexts, we  show that MLDG-Seg is able to learn generalized representations from very \textit{limited training examples}. Finally, we show that the learned generalized representation can be quickly adapted with a few examples from the unseen target distribution in a \textit{k}-shot learning setting to achieve additional performance gains.
\section{Methodology}
Meta-learning, or \textit{learning to learn} \cite{maml}, aims at learning a variety of tasks, and then quickly adapting to new tasks in different settings. We adopt the optimization-based model-agnostic method proposed in \cite{MLDG}, called Meta Learning Domain Generalization (MLDG). Here, we briefly describe the method.

\begin{SCfigure}
    \includegraphics[height=1.8in, width=2.7in]{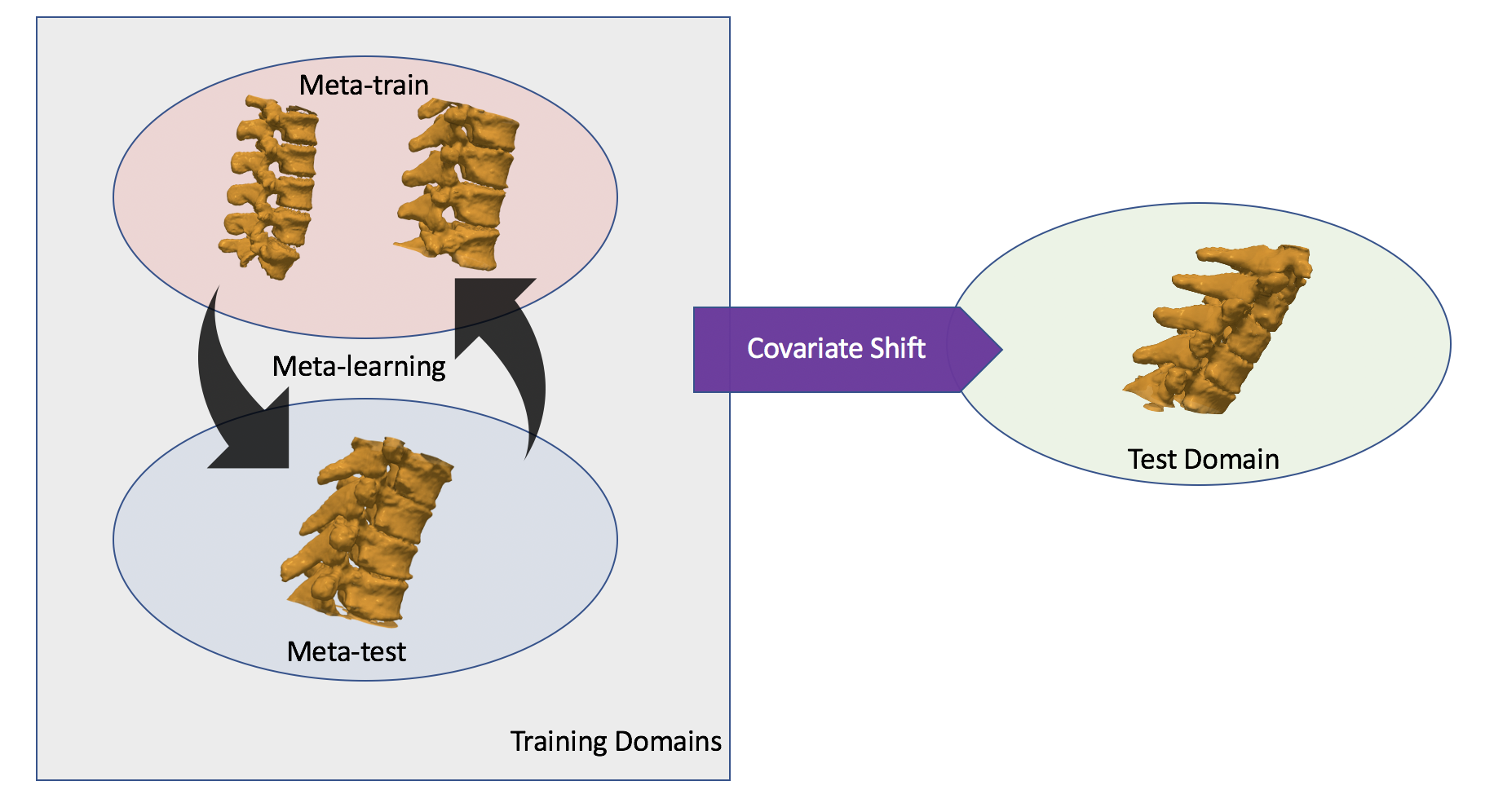}
    \caption{\textbf{MLDG-Seg}: A schematic. Our method extends the MLDG algorithm to a segmentation by using a 3D Unet-like architecture as the backbone. To explain the segmentation procedure, consider the four domains: lumbar ($\mathcal{D}$1), lower ($\mathcal{D}$2), middle ($\mathcal{D}$3), upper ($\mathcal{D}$4) thoracic regions. As an example, the domains $\mathcal{D}$1, and $\mathcal{D}$2 comprise the meta-train set, and $\mathcal{D}$3 the meta-test. The domain $\mathcal{D}$4 is the held-out test domain.}
    \label{MLDG-Seg}
\end{SCfigure}

\textit{\textbf{Description.}} Let there be two distributions: source $\mathcal{S}$, and target $\mathcal{T}$. Both $\mathcal{S}$, and $\mathcal{T}$ share the same task, for example, segmentation or classification with the same label space. The goal of MLDG is to learn a single set of model parameters $\mathcal{\theta}$ via gradient descent and two meta-learners: meta-train and meta-test procedures. The model is trained on only the source $\mathcal{S}$ domains, and then tested on target $\mathcal{T}$ domain. The source domains $\mathcal{S}$ are split into two sets: meta-train domains $\hat{S}$, and meta-test domains $\bar{S}$ = $\mathcal{S}$ - $\hat{S}$. The goal of the two splits is to mimic the setting of domain shifts, and thereby make it easier for the model to generalize on an unseen target domain $\mathcal{T}$. We reproduce the learning procedure in Algorithm 1, and show the extended version in Fig. \ref{MLDG-Seg}.
\begin{algorithm}[t!]
\label{algorithm_MLDG}
	\caption{MLDG}
	\begin{algorithmic}[1]
	\State \textbf{Input: Source domains $\mathcal{S}$}
	\State Model parameters $\mathcal{\theta}$ and Hyperparameters: $\alpha$, $\beta$, $\gamma$.
		\For {$iterations=1,2,\ldots$}
			\State \textbf{Randomly Split} source domains $\mathcal{S}$ into meta-train $\hat{S}$, and meta-test $\bar{S}$
			\State \textbf{Meta-train}: Gradients $\nabla_{\theta} = F^{'}_{\theta}(\hat{S}; \theta)$
			\State Updated parameters: $\theta^{'} {\leftarrow} \theta - \alpha \nabla_{\theta}$
			\State \textbf{Meta-test}: Compute Loss with updated parameter $\theta^{'}$ as $G(\bar{S}; \theta^{'})$
		    \State \textbf{Final Model parameters}: $\theta {\leftarrow} \theta - \gamma \frac{\partial (F(\hat{S}; \theta)+ \beta G(\bar{S}; \theta - \alpha \nabla_{\theta}))}{\partial \theta}$
		\EndFor
	\end{algorithmic}
\end{algorithm}

\textit{\textbf{Explanation.}} Let's consider a motivating example (carried out later in Experiment 1) from the image manifold of lumbar ($\mathcal{D}$1), lower thoracic ($\mathcal{D}$2) and middle thoracic regions ($\mathcal{D}$3), comprising the set of source domains $\mathcal{S}$; as well as upper ($\mathcal{D}$4) thoracic region, which constitutes the unseen test $\mathcal{T}$ domain. Refer to Algorithm 1. The MLDG method is supposed to learn a single model parameter $\mathcal{\theta}$ with the help of two optimization steps. At every iteration, the set of images in the source domains, here ($\mathcal{D}$1, $\mathcal{D}$2, and $\mathcal{D}$3) are randomly split into a meta-train (for example, consisting of images from $\mathcal{D}$1, $\mathcal{D}$2 set), and meta-test with the set $\mathcal{D}$3. Now, two losses are computed. The first loss $\mathcal{F}$ is computed using the training examples from meta-train set and the gradient is computed with respect to the model parameter $\theta$. The second loss $\mathcal{G}$ is computed on the meta-test set with the updated parameter $\theta^{'}$ = $\theta - \alpha \nabla_{\theta}$. The key idea by introducing the second loss in the meta-test stage is that an improvement of the model's performance on the meta-train set should also improve the model's performance on the meta-test set. The final model parameter $\mathcal{\theta}$ is updated by taking the gradient of the weighted combination of the two losses $\mathcal{F}$, and $\mathcal{G}$. By doing so, the model is tuned in such a way that performance is improved in both meta-train, and meta-test domains. In other words, the model is regularized and does not overfit to one particular domain, by finding the best possible gradient direction due to the joint optimization of the two losses. Compare this to a ``vanilla'' setup, where a model is directly given images from the three domains $\mathcal{D}$1, $\mathcal{D}$2, and $\mathcal{D}$3 without any meta-learning setup, might overfit to a domain by minimizing its loss, and maximizing the loss for the other domains.
\section{Experiments}
\subsection{Databases}
We validate our approach on three publicly available CT vertebrae segmentation datasets. See supplementary material for sample images of the three datasets.

\textbf{CSI challenge - healthy cases.} The datasets of spine CT (MICCAI 2014 challenge \cite{csi}) were acquired during daily clinical routine work in a trauma center from 10 adults (age: 16 to 35 years). In each subject, all 12 thoracic and 5 lumbar vertebrae were manually segmented to obtain ground truth.

\textbf{xVertSeg segmentation challenge - pathology cases.}
This database consists of fractured and non-fractured CT lumbar spine images. We used the 15 subjects, ranging in age from 40 to 90 years, made publicly available with their corresponding lumbar ground truth segmentations \cite{xvertseg}.

\textbf{VerSe MICCAI segmentation challenge 2020 - versatile dataset.}
This database consists of labeled lumbar, thoracic, cervical vertebrae across 100 cases in the released set \cite{verse}, \cite{verse_radiology} as of June 2020. The data comes from several medical centers, with a very wide range of acquisition protocols, certain vertebrae with surgical implants, and a range of FoV with arbitrary image orientations.
\subsection{Experimental setup}
\label{experimental_setup}
We define the different implemented procedures, and then perform three experiments.
\textbf{Baseline}
is the vanilla 3D Unet-like \cite{unet} architecture. The model is trained on the source domains, and then tested on the unseen held-out test domain. The procedure is then repeated for the other domains.
\textbf{MLDG-Seg}
is the vanilla 3D Unet-like architecture trained under the scheme in Algorithm 1. The model is trained on the source domains, but at every iteration the source domains are divided into meta-train and meta-test splits. The model is then tested on unseen held-out test domain. The procedure is then repeated for the other domains.
\textbf{\textit{k}-shot learning}
Once the MLDG-Seg model has been trained on source domains, it can be quickly adapted via fine-tuning using very limited labeled examples from the unseen target domain. The weights of the encoder, and bottleneck layers of the 3D Unet-like architecture were frozen in the \textit{k}-shot learning experiments for fine-tuning. Our approach of \textit{k}-shot learning is different than that of \textit{test time adaptation} methods \cite{test_time_1, test_time_2, test_time_3, test_time_4}, where usually the encoder weights are not frozen in a Y-shaped network architecture, but are explicitly updated using a few examples at the time of testing. In contrast, we fine-tune the decoder weights with \textit{k} examples from the test distribution.
\textbf{Oracle}
To establish a theoretical upper bound on segmentation accuracy, we train the vanilla 3D Unet-like architecture using labeled examples from the target domain, and then test on held-out examples from the target domain. This is presumably the easiest task, since the training and test domain distribution are similar. However, in our relatively small datasets the amount of available data for the oracle experiment is smaller than for the domain generalization experiments, so oracle performance reported below should be read with caution.

\textbf{Experiment 1.} The images in all the subjects of the CSI database are divided into different regions: lumbar (L1-L5), lower thoracic (T9-T12), middle thoracic (T5-T8), and upper thoracic vertebrae (T1-T4). Each of these regions comprise of a \textbf{domain}. Here, the aim is to let the model generalize across the underlying vertebral anatomy, the intensity profile and the surrounding intervertebral disc space which varies significantly along the vertebral column.

\textbf{Experiment 2.} Here, the four domains i.e., lumbar, lower, middle, and upper thoracic vertebrae from the CSI healthy database comprise the set of source domains. The xVertSeg database with pathology cases comprise the unseen target domain. Here, the aim is to let the model generalize from the vertebrae of healthy subjects to the structure of fractured or dislocated vertebrae images obtained at a different medical center.

\textbf{Experiment 3.} Here, the model is trained on CSI and xVertSeg datasets and tested on VerSe dataset, with the aim to generalize to unseen anatomies, different acquisition protocols, arbitrary orientations, FoVs, and pathology.

\textbf{Train, Validation, and Test data split-up details.} The supplementary material details the training, validation, and test sets for all three databases.

\textbf{Implementation details.}
All the images were converted to 1 mm$^{3}$ isotropic resolution using FreeSurfer \cite{freesurfer}. The images were standardized, using mean subtraction and division by standard deviation, and then normalized between 0 and 1. We implemented a 3D Unet-like \cite{unet} architecture in PyTorch \cite{pytorch}. This 3D Unet is the backbone architecture used for all the experiments. Each of the three hyperparameters $\alpha$, $\beta$, and $\gamma$ in MLDG (Algorithm 1) are set to 1. We use stochastic gradient descent as the optimizer with a learning rate of 0.001, momentum of 0.9, and a weight decay of 5 x 10$^{-5}$, dropout with probability of 0.3, and \textit{groupnorm} as the normalization technique. See supplementary material for more details on the network architecture. The loss function used is Generalized Dice Loss \cite{dice}. We randomly sample 50 patches of size 64x64x64 from each subject in a domain, and perform data augmentation by randomly rotating and flipping 30 patches out of these 50 patches for every procedure. For the \textit{k}-shot procedure, we randomly sampled 5 patches from each of the \textit{k}-th subject from the unseen distribution. We trained every procedure, as described above, in experiment 1 for 10 epochs, and experiments 2 and 3 for 15 epochs.  We use the model that gave the highest Dice score on the validation set to evaluate the unseen test domain. The models were trained on Nvidia P100 Tesla GPUs. A sliding window approach was used to obtain the predicted segmentations, which were then post-processed by retaining the largest connected component.

\textbf{Evaluation details.} We compute Dice coefficient (\%), a volume-based measurement, and Average Symmetric Surface Distance (ASSD) in mm, a surface-based metric between the groundtruth image and a given model procedure segmentation output. \textit{Desired: Higher Dice, and lower ASSD scores.} Furthermore, we perform pairwise Wilcoxon signed-rank test \cite{wilcoxon} between the baseline and each of the other procedures in all the three experiments for both Dice score and ASSD. In the three Tables \ref{Exp1Table}, \ref{Exp2Table} and \ref{Exp3Table}, we highlight the procedures which reach significance at an $\alpha$=0.05 significance value using the following notation to denote the level of significance: * ($p<$0.05), ** ($p<$0.005), and *** ($p<$0.0005).
\section{Results and Discussion}
\begin{SCfigure}
    \includegraphics[height=2.6in, width=3.0in]{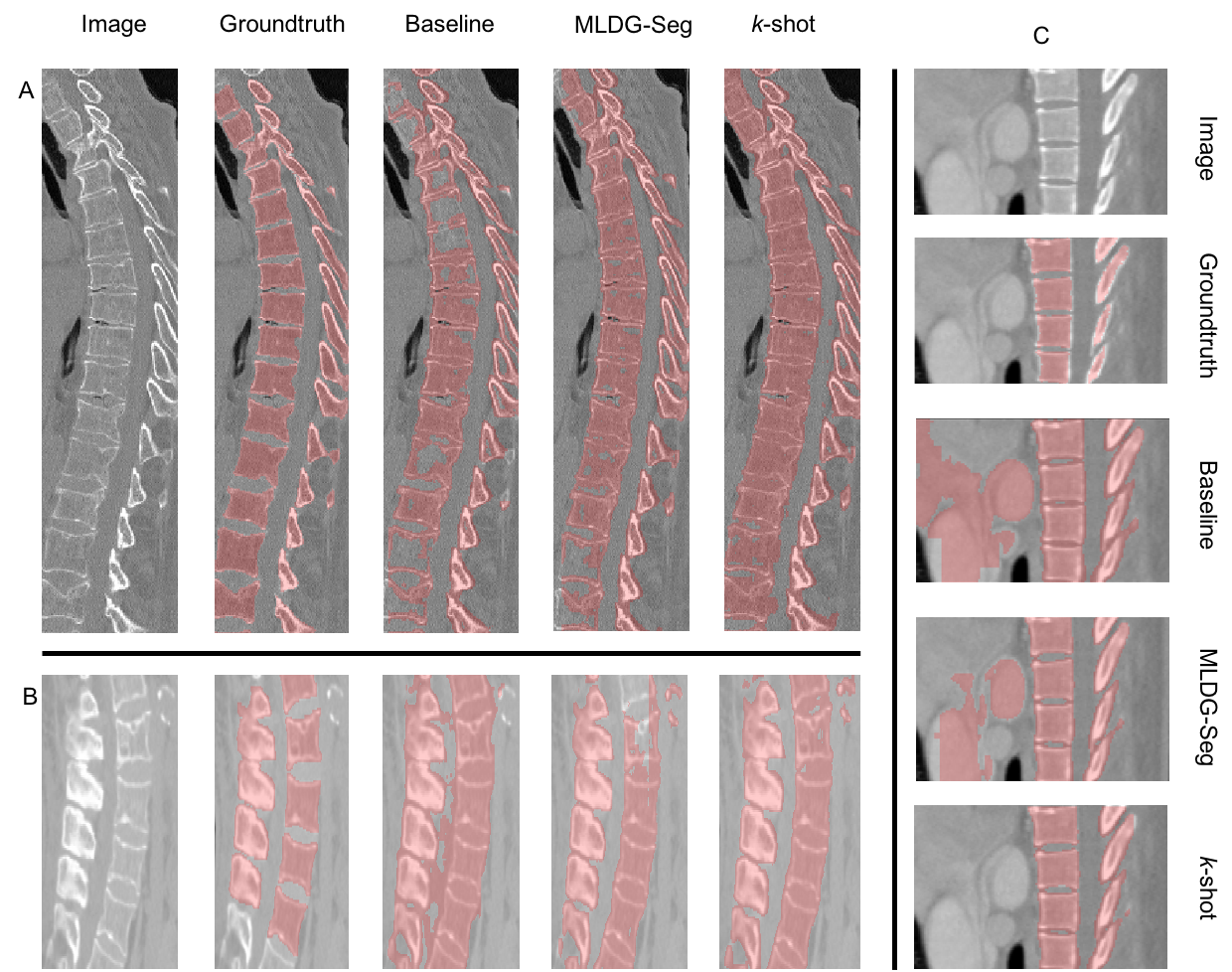}
    \caption{Qualitative illustrations for the three experiments. (A) Experiment 3: tested on VerSe dataset, here: \textit{k}=7. (B) Experiment 2: tested on xVert dataset, here: \textit{k}=4. (C) Experiment 1: shown is the middle thoracic region as the test set, here: \textit{k}=2. Qualitative results for lumbar, lower, and upper thoracic regions can be found in the supplement. A minor discrepancy at the boundaries is noticed in groundtruth in C, perhaps due to registration errors.}
\label{output_figures}
\end{SCfigure}

\textbf{Experiment 1.}
Table \ref{Exp1Table} tabulates the results on different held-out test distributions. Each model is trained on three domains and then tested on the fourth unseen domain, except the oracle which is trained and tested on the same domain. MLDG-Seg consistently outperforms the baseline on both Dice and ASSD, and shows the desired low variance amongst the subjects. With a very few labeled examples from the test distribution, we see a further boost in performance. Fig. \ref{output_figures}C shows that the MLDG-Seg is better able to segment the region of interest (ROI) by having significantly reduced background error than the baseline, in addition to the correct delineation of intervertebral discs (IVDs) when the held-out distribution is the middle thoracic region. A further improvement in performance is obtained using additional \textit{k}-shot examples over MLDG-Seg. Ideally, the oracle should have the best performance than the rest of the procedures as the test domain distribution is similar to the training domain distribution. Here in this particular experimental setup, it is \textit{not} surprising to see that the MLDG-Seg (and in some test domains, the baseline) performs better than the oracle. This might be due to the limited number of training (and validation) examples available to train the oracle. Furthermore, since all the procedures were trained for a fixed number of epochs, the oracle might have had a disadvantage of being trained for a lesser number of gradient steps than the MLDG-Seg procedure. An alternative approach would be to train a single oracle model which learns from all the domains simultaneously, and then test on each domain separately.

\begin{figure}[h!]
    \centering
    \subfigure{%
    \includegraphics[height=1.5in, width=2.2in]{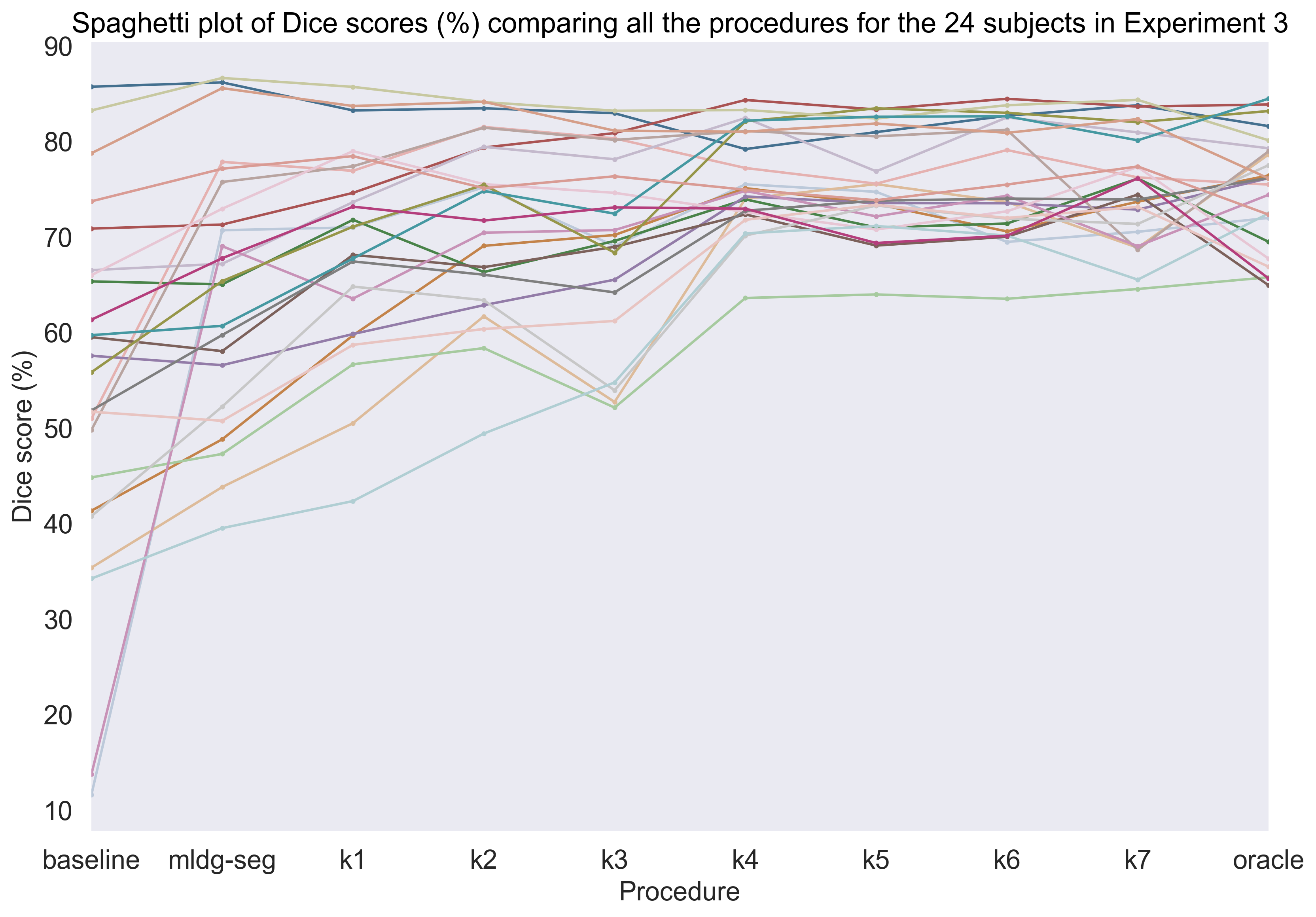}}%
    \qquad
    \subfigure{%
    \includegraphics[height=1.5in, width=2.2in]{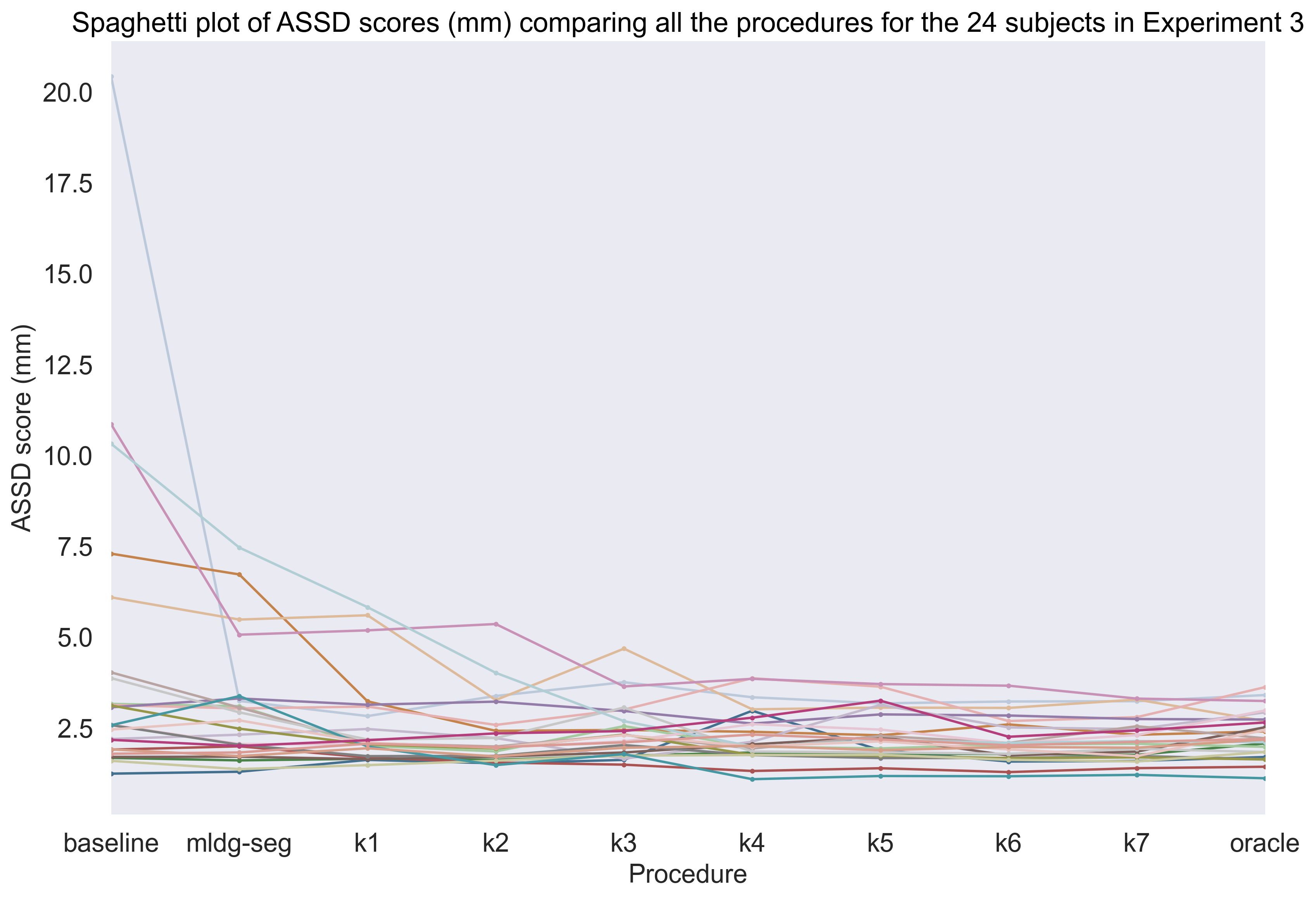}}%
    \caption{Spaghetti plots are shown for Experiment 3. \textit{Left}: Dice coefficient, \textit{Right}: ASSD score. Y-axis shows the Dice (\%), and the ASSD (mm) scores for the different procedures shown on the X-axis. Each of the plotted lines denote a subject. Therefore, one can track the performance of a procedure for a given subject. See supplementary material for the spaghetti plots for all the other experiments.}
    \label{spaghetti_exp3}
\end{figure}

\textbf{Experiment 2.}
  Table \ref{Exp2Table} and the Spaghetti plots in Fig. \ref{spaghetti_exp3} shows that the MLDG-Seg outperforms the baseline. Fig. \ref{output_figures}B shows that MLDG-Seg is able to delineate the vertebrae by \textit{not} segmenting the undesired spinal cord, as incorrectly segmented by the baseline. The \textit{k}-shot setting further improves the segmentation. Here again, it is \textit{not} necessarily surprising to see that MLDG-Seg performs better than the oracle. The oracle was trained with a smaller number of subjects than the baseline and MDG-Seg (see supplement).

\textbf{Experiment 3.}
Table \ref{Exp3Table} shows the improved performance of MLDG-Seg over the baseline, and consistent performance boost in the few-shot learning regime. The \textit{k}-shot procedure is either on par with the oracle or outperforms the same for \textit{k}=4, 5 and 6. Fig. \ref{output_figures}A depicts the superior performance of MLDG-Seg over the baseline on a compression fraction subject from a different distribution than the training set, where MLDG-Seg is able to segment more vertebrae completely than the baseline. The \textit{k}-shot setting further improves the performance by segmenting the remaining vertebrae not segmented by baseline or MLDG-Seg. Here, the oracle performs better than most of the procedures, which is due to the fact that the training set for oracle is relatively larger than in experiments 1 and 2.\\

\begin{table}[t!]
  \caption{Dice score (\%), and ASSD in mm (mean $\pm$ std. dev) for Experiment 1. Each row reports the result on held-out unseen test domain, where the model was trained on the remaining three domains. Supplement contains results on additional subjects.}
  \centering
    \scalebox{0.90}
    {\begin{tabular}{b|b|a|b|b|b}
    Test domain & \makecell{Baseline} & \thead{MLDG-Seg}  & \makecell{\textit{k}=1} & \makecell{\textit{k}=2} & \makecell{Oracle}\\
    \midrule
    \hline
    Lumbar & 
    81.67 $\pm$ 8.45 (1.85 $\pm$ 0.75) & 
    87.85 $\pm$ 2.73 (1.42 $\pm$ 0.31) & 
    88.57 $\pm$ 1.53 (1.46 $\pm$ 0.24) &
    88.19 $\pm$ 2.47 (1.69 $\pm$ 0.67) &
    83.60 $\pm$ 2.68 (2.74 $\pm$ 0.37)
    \\
    \hline
    Lower Th &
    83.52 $\pm$ 4.12 (2.66 $\pm$ 0.98)&
    86.17 $\pm$ 2.17 (1.44 $\pm$ 0.09)&
    81.44 $\pm$ 2.46 (2.73 $\pm$ 0.49)&
    82.28 $\pm$ 1.49 (2.74 $\pm$ 0.55)&
    80.25 $\pm$ 4.16 (2.32 $\pm$ 0.41)\\
    \hline
    Middle Th &
    55.72 $\pm$ 4.13 (10.41 $\pm$ 0.42)&
    64.36 $\pm$ 11.45 (6.85 $\pm$ 3.18)&
    75.57 $\pm$ 5.60 (3.56 $\pm$ 1.64)&
    76.98 $\pm$ 8.66 (2.20 $\pm$ 1.45)&
    83.60 $\pm$ 0.58 (2.21 $\pm$ 0.54)\\
    \hline
    Upper Th &
    82.00 $\pm$ 1.45 (1.68 $\pm$ 0.18) &
    83.70 $\pm$ 2.19 (1.46 $\pm$ 0.20) &
    74.47 $\pm$ 6.21 (4.83 $\pm$ 2.56) &
    81.50 $\pm$ 4.07 (1.75 $\pm$ 0.39) &
    75.84 $\pm$ 4.00 (1.93 $\pm$ 0.36)
  \end{tabular}}
  \label{Exp1Table}
\end{table}

    \begin{table}[h!]
      \caption{Dice score (\%), and ASSD in mm (mean $\pm$ std. dev)  for Experiment 2. The model was trained on CSI dataset and tested on xVertSeg dataset, thereby generalizing to a pathology dataset when trained on a healthy population.}
      \centering 
      \scalebox{0.75}
        {\begin{tabular}{f|e|f|f|f|f|f}
        \makecell{Baseline} & \thead{MLDG-Seg} & \makecell{\textit{k}=1} & \makecell{\textit{k}=2} & \makecell{\textit{k}=3} & \makecell{\textit{k}=4} & \makecell{Oracle}\\
        \midrule
        \hline
        74.13 $\pm$ 13.69 (3.16 $\pm$ 1.05) & 
        75.34 $\pm$ 12.10 (2.42 $\pm$ 0.70$^{**}$) & 
        76.88 $\pm$ 7.35 (2.74 $\pm$ 0.94$^{*}$) & 
        76.51 $\pm$ 8.85 (2.31 $\pm$ 0.65$^{**}$) & 
        77.95 $\pm$ 8.79$^{*}$ (2.38 $\pm$ 0.74$^{**}$) & 
        79.20 $\pm$ 8.01$^{*}$ (2.29 $\pm$ 0.77$^{**}$) &
        74.94 $\pm$ 7.84 (3.98 $\pm$ 1.62)
    \end{tabular}}
  \label{Exp2Table}
\end{table}
\begin{SCtable}
  \caption{Dice score (\%), and ASSD in mm (mean $\pm$ std. dev) for Experiment 3. The model was trained on CSI, and xVertSeg dataset and tested on VerSe dataset, thereby generalizing to fractured or dislocated vertebrae images obtained at different medical centers, unseen anatomies, different acquisition protocols, arbitrary orientations, FoVs, and various pathology.}
    \scalebox{0.8}
    {\begin{tabular}{d|d|d}
    \makecell{Procedure} & \makecell{Dice score} & \makecell{ASSD}\\
    \midrule
    \hline
    Baseline & 54.58 $\pm$ 18.16 & 4.23 $\pm$ 4.21\\
    \rowcolor{lightgray} \textbf{MLDG-Seg} & 64.82 $\pm$ 13.13$^{***}$ & 2.99 $\pm$ 1.59$^{*}$ \\
    \textit{k}=1 & 69.12 $\pm$ 10.48$^{***}$ & 2.57 $\pm$ 1.21$^{**}$\\
    \textit{k}=2 & 71.49 $\pm$ 9.04$^{***}$  & 2.31 $\pm$ 0.91$^{***}$\\
    \textit{k}=3 & 70.18 $\pm$ 9.53$^{***}$  & 2.42 $\pm$ 0.76$^{**}$\\
    \textit{k}=4 & 75.90 $\pm$ 4.99$^{***}$ & 2.29 $\pm$ 0.69$^{*}$\\
    \textit{k}=5 & 75.27 $\pm$ 5.11$^{***}$  & 2.31 $\pm$ 0.68$^{*}$\\
    \textit{k}=6 & 75.54 $\pm$ 5.71$^{***}$  & 2.13 $\pm$ 0.60$^{**}$\\
    \textit{k}=7 & 75.28 $\pm$ 5.67$^{***}$  & 2.17 $\pm$ 0.57$^{**}$\\
    Oracle       & 74.97 $\pm$ 5.81$^{***}$ & 2.31 $\pm$ 0.62$^{*}$
  \end{tabular}}
  \label{Exp3Table}
\end{SCtable}
\textbf{Conclusion}
In the present work, we benchmarked the performance of a gradient-based meta-learning domain generalization segmentation method in the context of biomedical image analysis, across a variety of training and test settings. The method was not only able to generalize across multiple medical sites and scanners, which is the most widely studied problem of generalization, but was also able to generalize to newly introduced settings of unseen complex vertebrae anatomies, surrounding inter-vertebral discs space, varying bone and soft tissue intensities distribution, diseased populations, different acquisition protocols, arbitrary orientations, and FoVs, thus resembling actual clinical settings. In future, we will evaluate the method on other modalities such as MRI, and US and compare with the recently proposed domain generalization methods. Our source code, scripts, and dataset-split files are available at:\\ \url{https://github.com/Pulkit-Khandelwal/medical-mldg-seg}.

\textbf{Acknowledgments} This work was supported by NIH grant R01 EB017255.


\newpage

\title{Domain Generalizer: A Few-shot Meta Learning Framework for Domain Generalization in Medical Imaging [\textbf{Supplementary Material}]}
\titlerunning{MLDG-Seg [Supplementary Material]}
\author{Pulkit Khandelwal\inst{1,2} \Envelope \and
Paul Yushkevich\inst{2}}

\authorrunning{P. Khandelwal and P. Yushkevich}

\institute{Department of Bioengineering, University of Pennsylvania, Philadelphia, PA, United States \and
Penn Image Computing and Science Laboratory, Department of Radiology, University of Pennsylvania, Philadelphia, PA, United States\\ \email{pulks@seas.upenn.edu}}
\maketitle

\section{Dataset split-up for train, validation, and test sets}
\begin{figure}[H]
\centering
\includegraphics[scale=0.4]{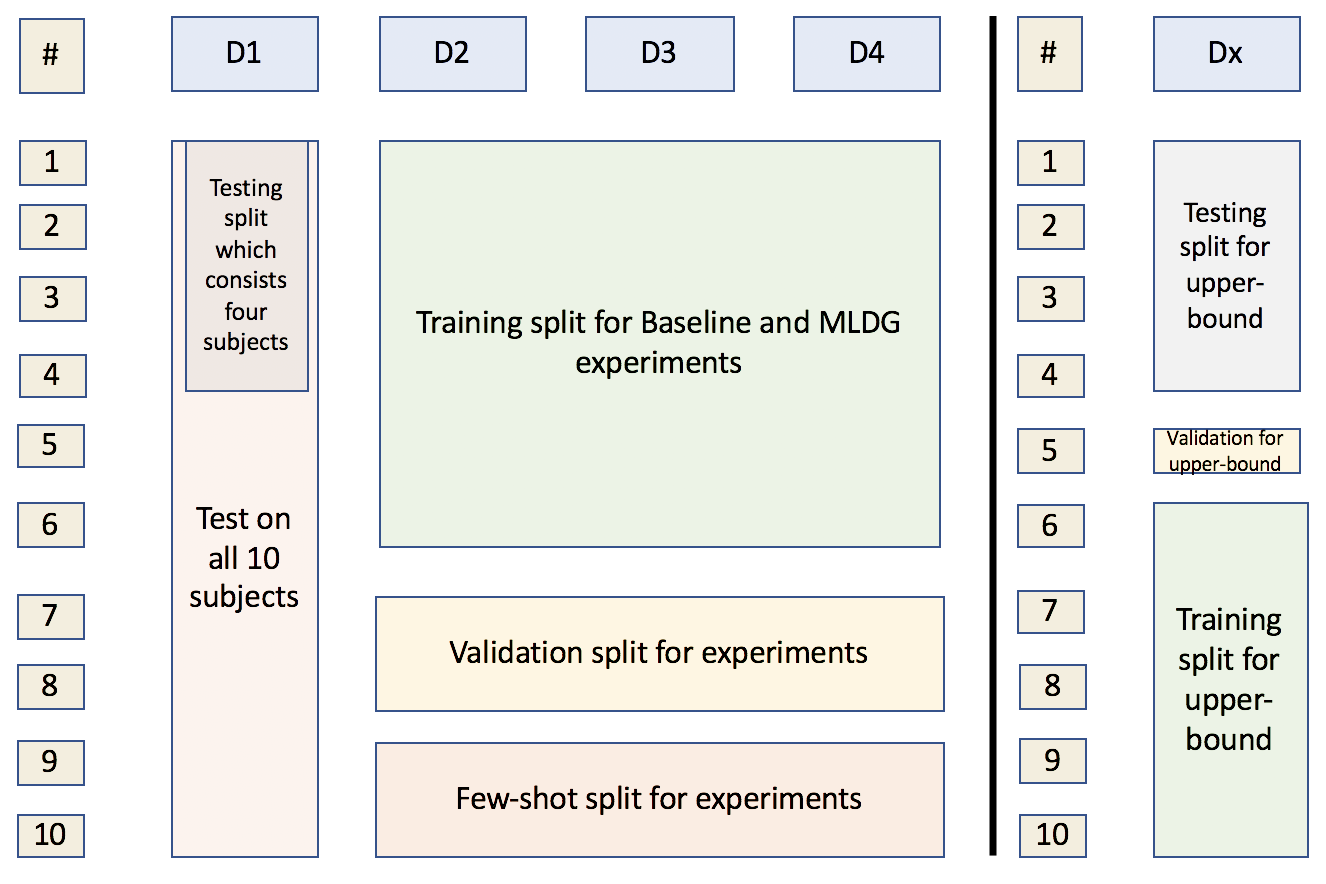}
\caption{The CSI database consists of 10 subjects each comprising the four domains: lumbar, lower, middle, and upper thoracic vertebrae. The numerals in the above figure denotes the subject number. The data split-up on the left side of the black vertical bar gives the split up for the results in Table 1 of the main paper. The data split for the oracle (upper-bound) procedure in Table 1 in the main paper is depicted on the right side of the vertical bar. The results in Table 1 in the main paper is for the first four subjects. The results are then reported for all the four domains, leaving one out as test. The results on all the 10 subjects are given in Table \ref{sec6_table} in Section \ref{last} of this supplementary material. Note: Dx denotes any of the domains.}
\label{split1}
\end{figure}

\begin{figure}[H]
\centering
\includegraphics[scale=0.6]{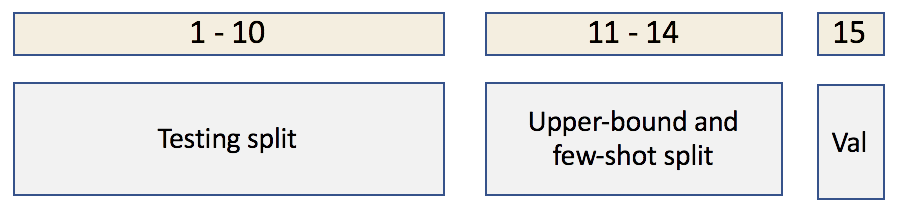}
\caption{The xVertSeg database consists of 15 subjects. Depicted above is the split for the results in Table 2 in the main paper. Note that oracle is the upper-bound in the figure.}
\label{split2}
\end{figure}

\begin{figure}[H]
\centering
\includegraphics[height=4.0in, width=4.7in]{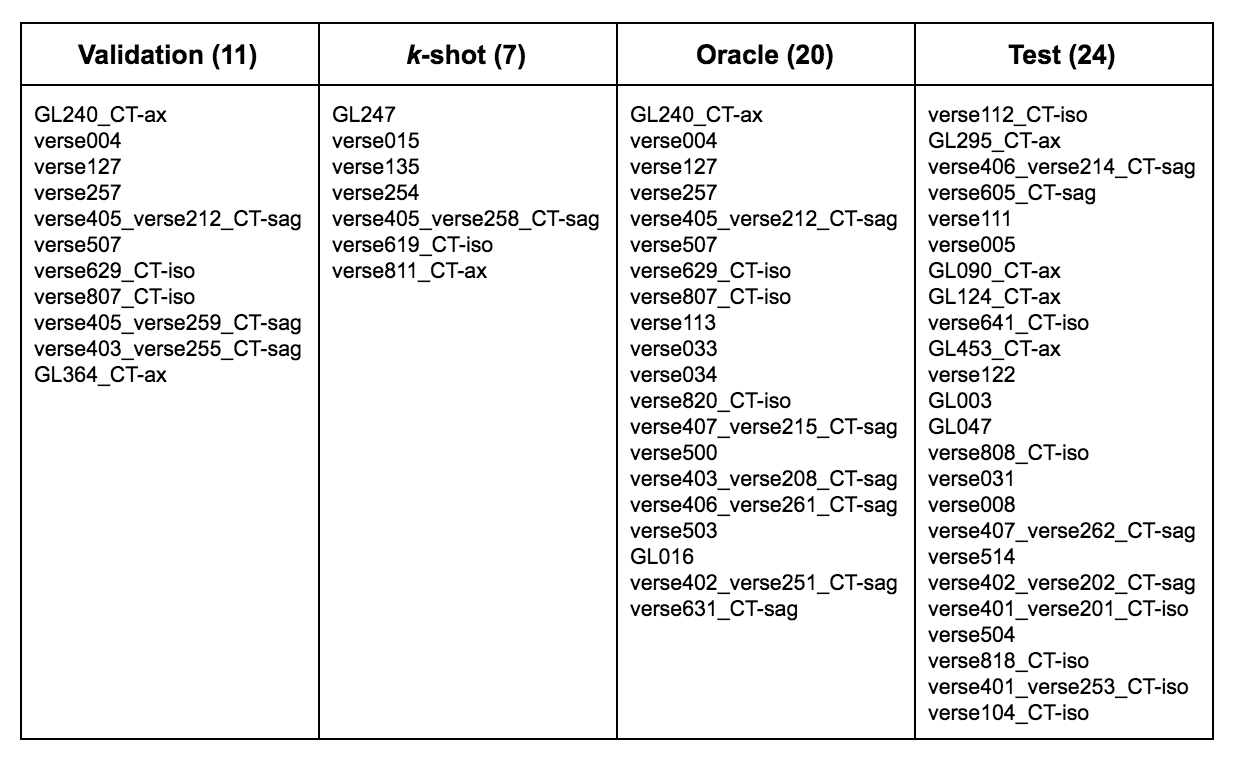}
\caption{The VerSe segmentation challenge dataset consists of 100 cases at the time of writing this manuscript in June 2020. We randomly select 55 cases, and assign them to the four categories as listed in the table. Shown are the subject IDs.}
\label{split3}
\end{figure}

\newpage
\section{3D Unet-like architecture}
\vspace{-6mm}
\begin{figure}[H]
\centering
\includegraphics[scale=0.40]{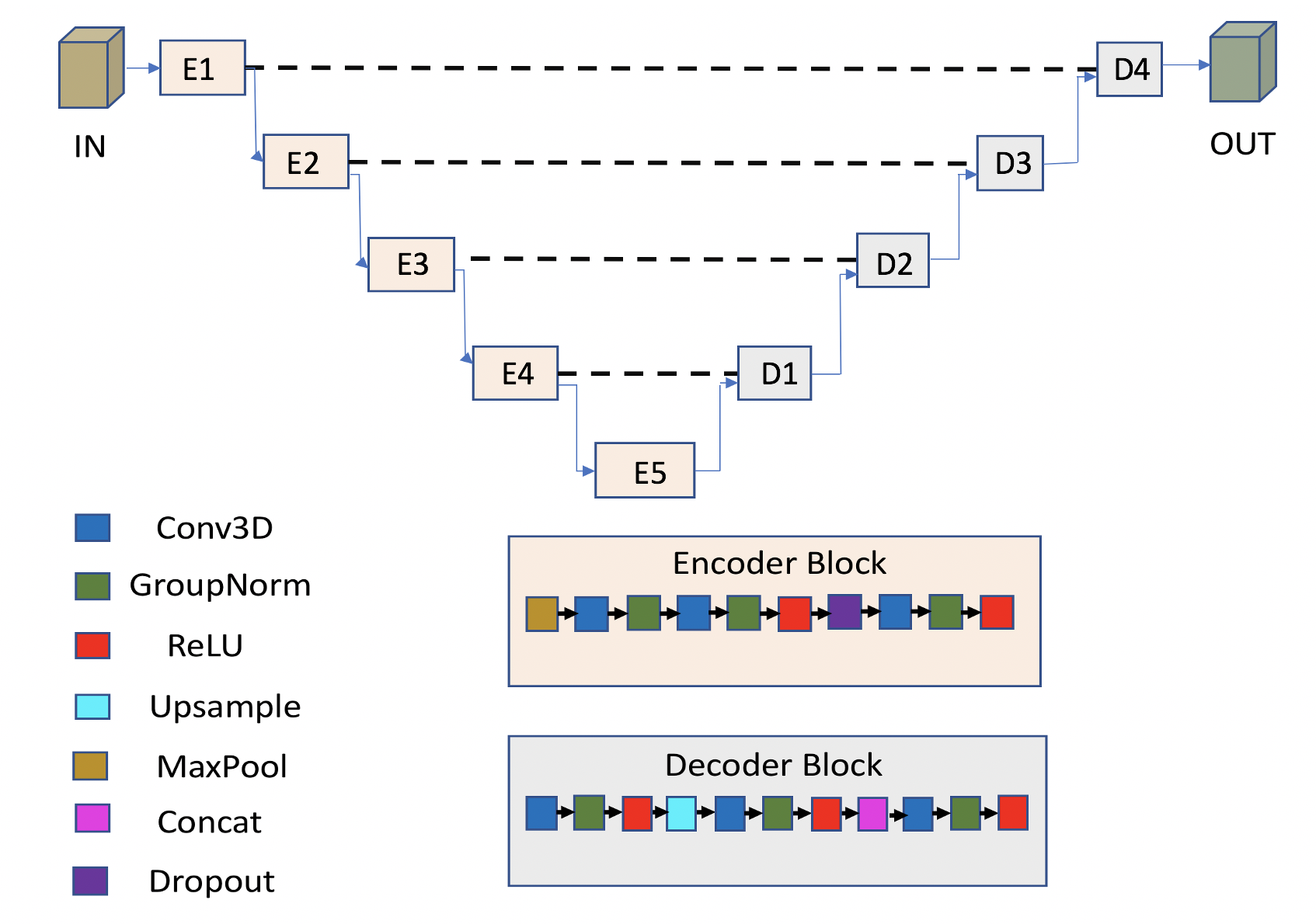}
\caption{3D Unet-like architecture implemented. IN and OUT are the input and the output respectively each with dimensions 1x64x64x64, and 2X64x64x64 (for the two classes). There are five encoder blocks (with the last block acting as the bottleneck layer). There are four decoder blocks. The dashed lines represent the skip connections. We have used dropout with probability of 0.3, and \emph{groupnorm} as the normalization method. Note: the first encoder block does not consist of the MaxPool operation, and the first decoder block does not consist of the first set of conv3D, groupnorm, and ReLU units. The \textit{\textbf{encoder}} block has the following \textit{output} dimensions at each of the encoder block: E1: 16x64x64x64, E2: 32x32x32x32, E3: 64x16x16x16, E4: 128x8x8x8, E5: 256x4x4x4. The \textit{\textbf{decoder}} block has the following \textit{output} dimensions at each of the decoder block: D1: 256x8x8x8, D2: 128x16x16x16, D3: 64x32x32x32, D4: 32x64x64x64.}
\label{unet}
\end{figure}

\newpage
\section{Sample image slices for the three datasets}
\begin{figure}[H]
    \centering
    \includegraphics[height=2.7in, width=3.5in]{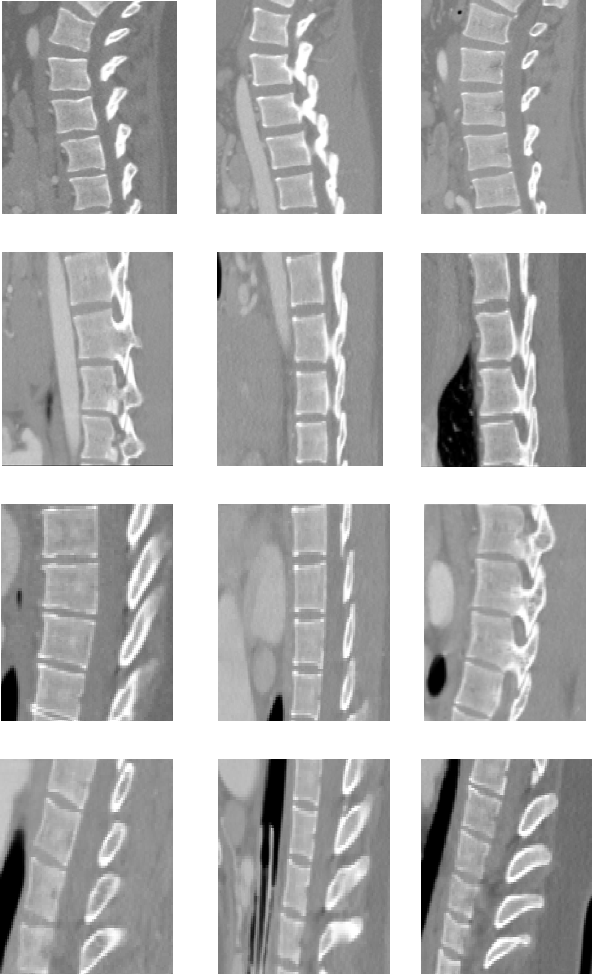}
    \caption{CSI dataset}
    \label{sample_slices}
\end{figure}

\begin{figure}[H]
    \centering
    \includegraphics[height=2.7in, width=3.5in]{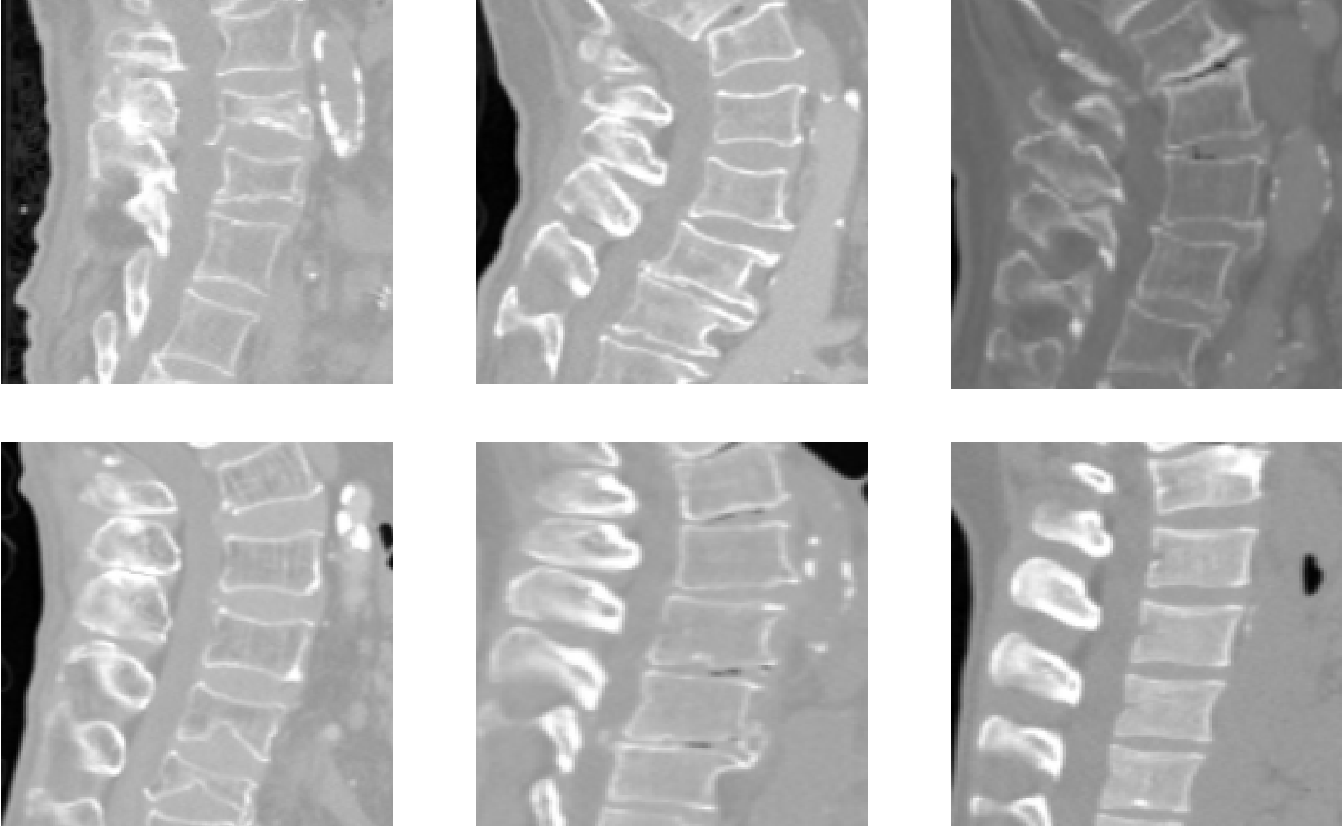}
    \caption{xVertSeg dataset}
    \label{sample_slices}
\end{figure}

\begin{figure}[H]
    \centering
    \includegraphics[height=3.0in, width=3.5in]{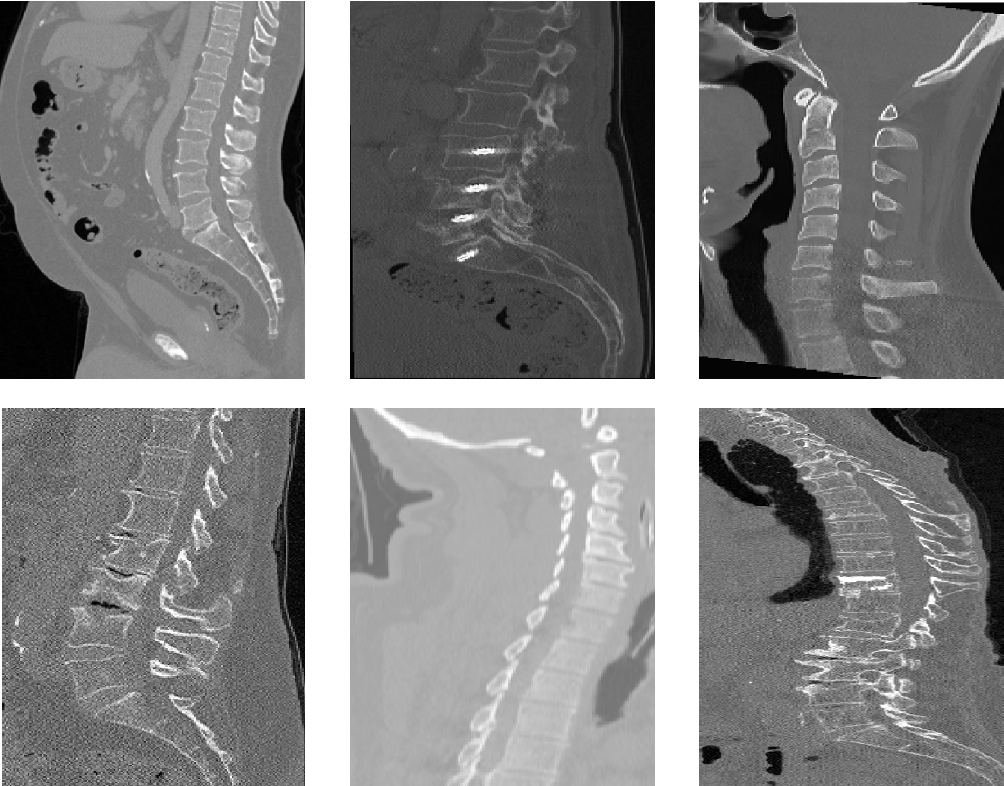}
    \caption{VerSe dataset}
    \label{sample_slices}
\end{figure}

\section{t-SNE plots}

\begin{figure}[H]
    \centering
    \includegraphics[height=2.0in, width=3.0in]{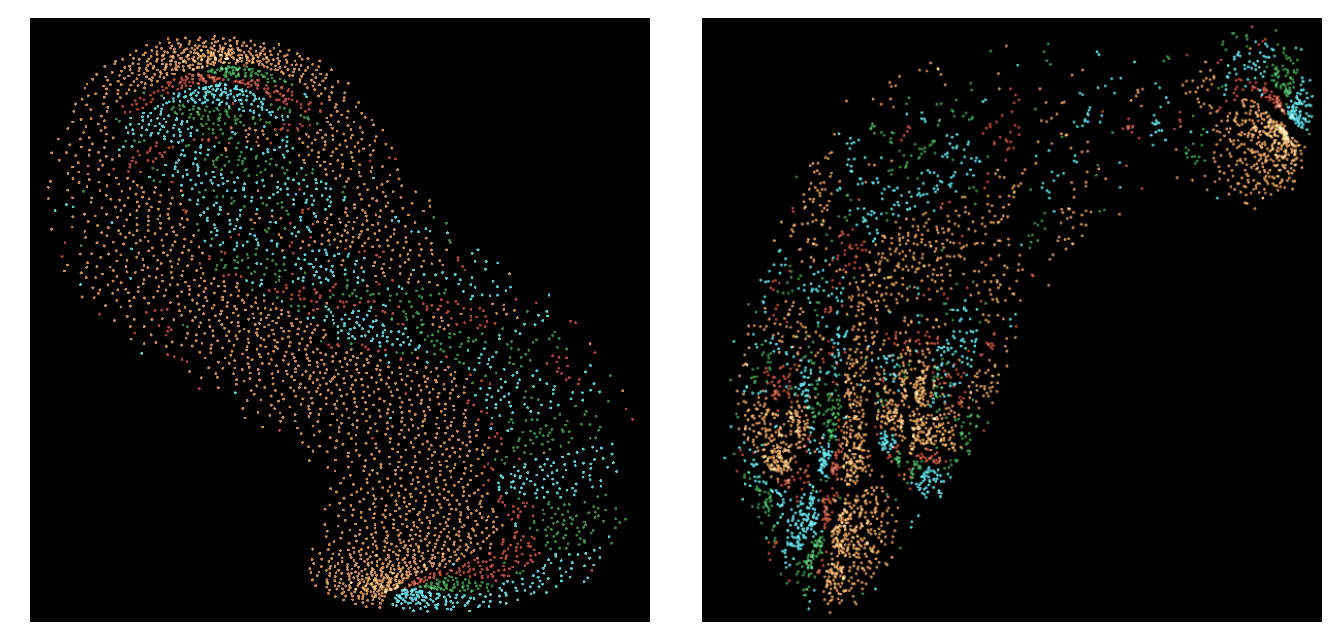}
    \caption{Shown are the t-SNE \cite{maaten2008visualizing} plots for Baseline (left), and MLDG-Seg (right). The trained bottleneck layer's feature vector with dimension of 16384 (E5: 256x4x4x4) are plotted for the experiment when the model was trained on CSI and xVertSeg dataset (Experiment 3). The four colors represent the four domains: lumbar, lower thoracic, middle thoracic, and upper thoracic regions in the CSI dataset. It can be seen that the data is clustered more separately in the trained baseline than the MLDG-Seg model. \textbf{Whereas, the trained MLDG-Seg method learns domain agnostic features to better generalize the model.} The following hyperparameters were used to produce the t-sne plot: perplexity: 45, learning rate: 10, supervision: 30, iterations: 700.}
\end{figure}

\section{Experiment 1 additional results}
\label{last}

\begin{table}[H]
  \caption{Results for Experiment 1 on all the \textbf{10 subjects}. See Figure 1 for the data split-up. Dice score (\%), and Average Symmetric Surface Distance (ASSD) in mm (mean $\pm$ std. dev) for Experiment 1 within brackets. Each row reports the result on the held-out unseen test domain, where the model was trained on the remaining three domains. We performed the pairwise Wilcoxon signed-rank test between the baseline and MLDG-Seg for both Dice score and ASSD. We highlight the procedures which reach significance at an $\alpha$=0.05 significance value using the following notation to denote the level of significance: * ($p$$<$0.05), ** ($p$$<$0.005), and *** ($p$$<$0.0005).}
  \centering
    \begin{tabular}{c|c|c}
    Test domain & Baseline & \textbf{MLDG-Seg} \\
    \midrule
    \hline
    Lumbar & 81.22 $\pm$ 9.48 (1.90 $\pm$ 0.80) &
    88.31 $\pm$ 2.55$^{*}$ (1.39 $\pm$ 0.32$^{*}$) \\
    Lower Thoracic & 83.58 $\pm$ 6.71 (2.39 $\pm$ 1.21) &
    86.77 $\pm$ 3.35 (1.41 $\pm$ 0.57$^{*}$) \\
    Middle Thoracic & 65.5 $\pm$ 15.14 (7.49 $\pm$ 4.17) &
    71.22 $\pm$ 14.4 (4.93 $\pm$ 3.80$^{**}$) \\
    Upper Thoracic & 81.55 $\pm$ 3.99 (1.59 $\pm$ 0.33) &
    81.75 $\pm$ 5.87 (1.70 $\pm$ 0.88) \\
  \end{tabular}
  \label{sec6_table}
\end{table}

\section{Experiment 1 additional qualitative results}
\begin{figure}[H]
\centering
\includegraphics[height=3.5in, width=4.0in]{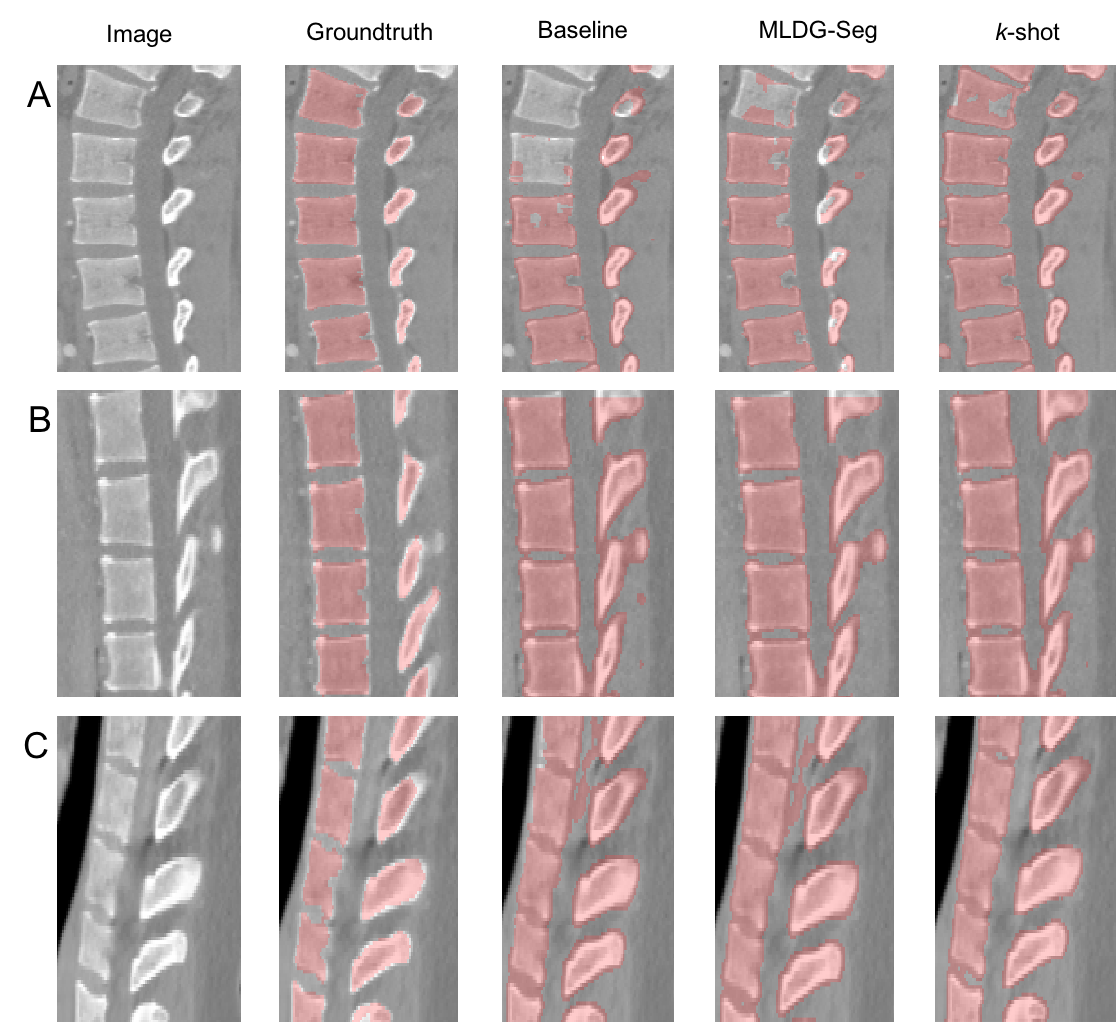}
\caption{Qualitative illustrations for the three held-out domains in experiment 1. (A) Held-out domain: Lumbar, here: \textit{k}=2. (B) Held-out domain: Lower Thoracic, here: \textit{k}=2. (C) Held-out domain: Middle Thoracic, here: \textit{k}=2. A minor discrepancy at the boundaries is noticed in the groundtruth segmentations, perhaps due to registration errors when converting to an isotropic resolution.}
\label{lowerTh}
\end{figure}

\newpage
\section{Spaghetti Plots}
\label{spaghetti}
In the following set of figures, Spaghetti plots are shown for all the experiments. Y-axis shows the Dice (\%), and the ASSD (mm) scores for the different procedures, shown on the X-axis. Each of the plotted lines denote a subject. Therefore, one can track the performance of a procedure for a given subject.

\subsection{Experiment 1 (4 subjects) [Refer Table 1 in the main paper]}
\label{spaghettiExp1_4}

\begin{figure}[H]
    \centering
    \subfigure{%
    \includegraphics[height=1.5in, width=2.2in]{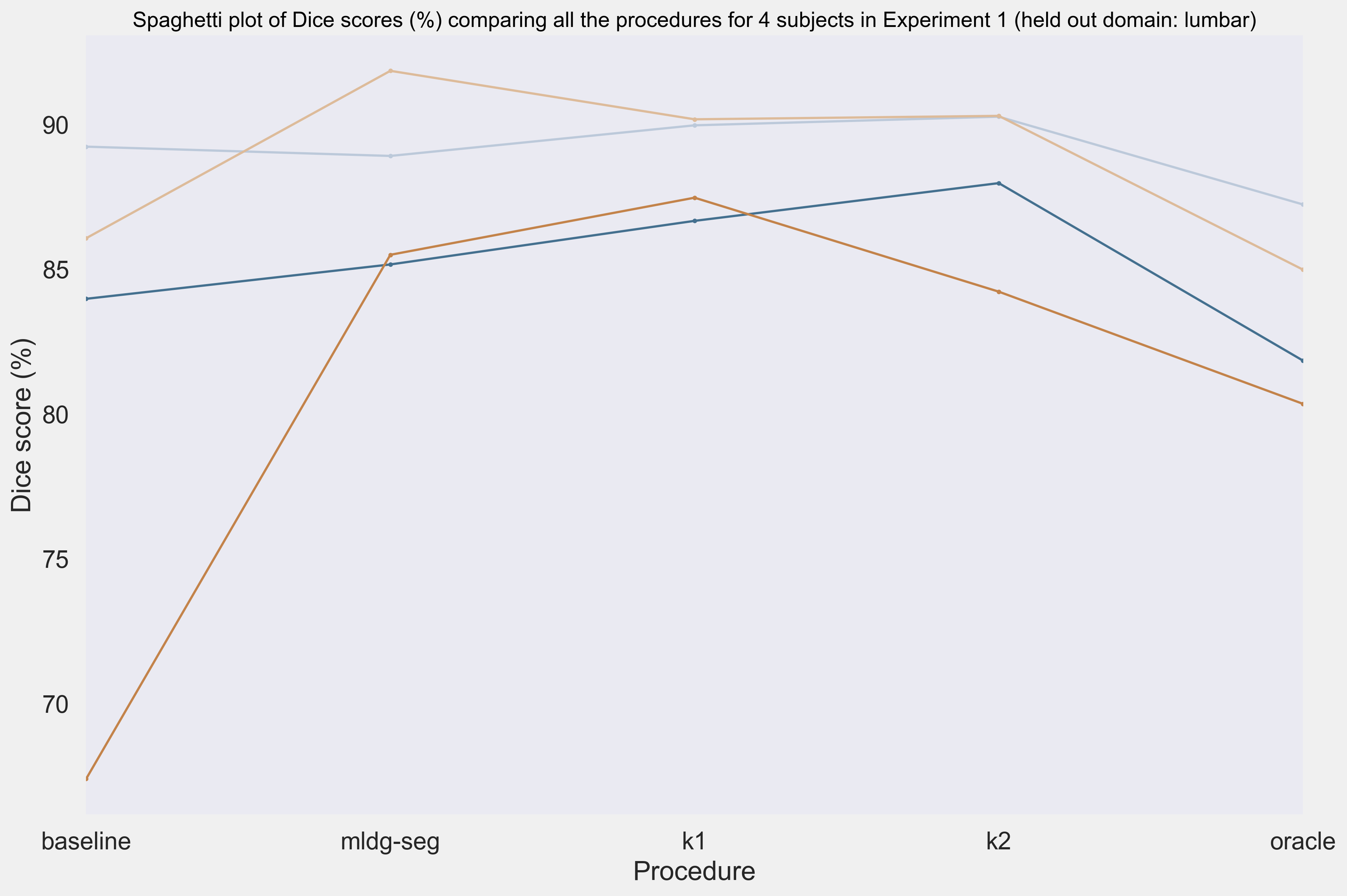}}%
    \qquad
    \subfigure{%
    \includegraphics[height=1.5in, width=2.2in]{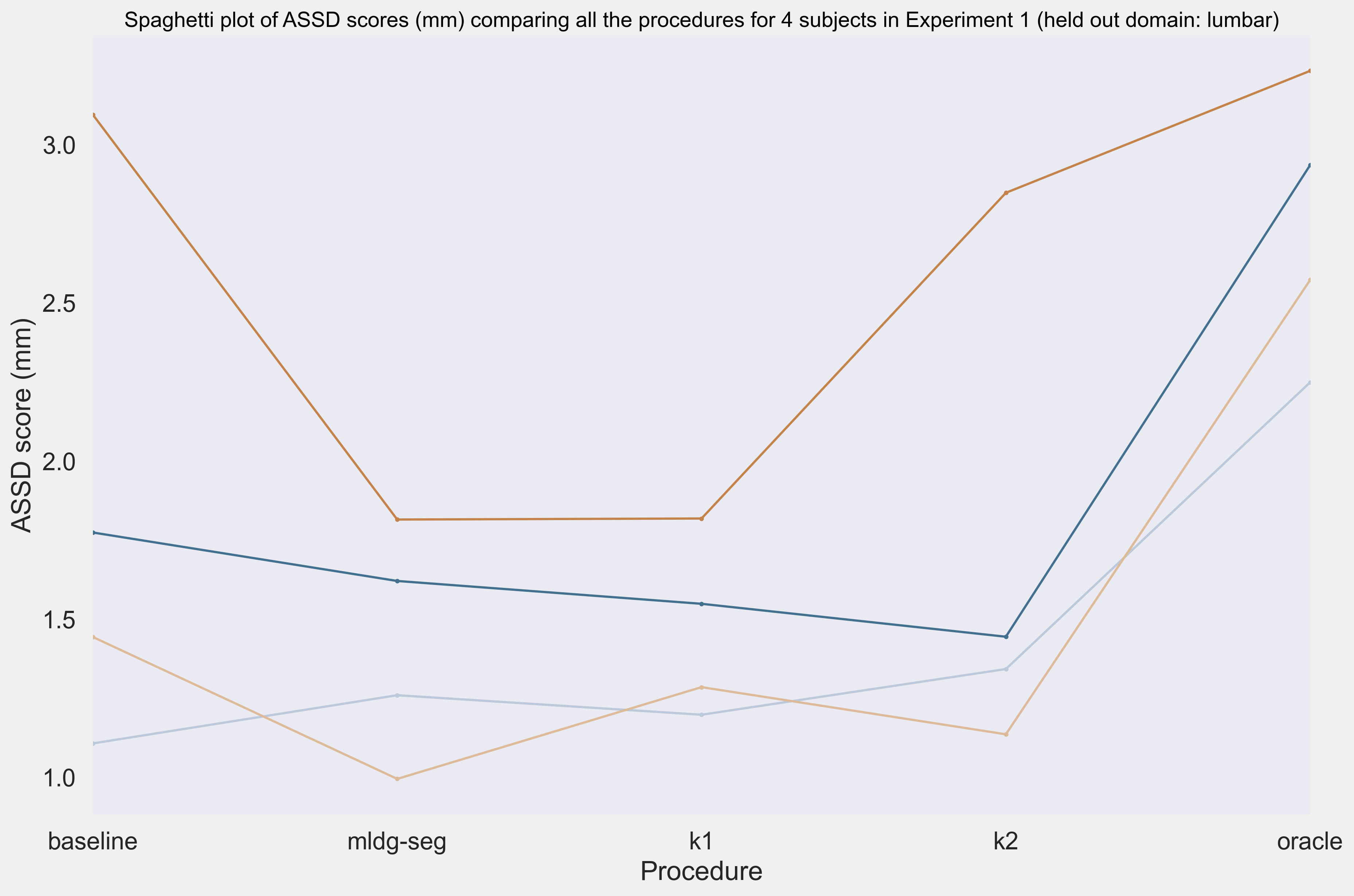}}%
    \caption{\textit{Left}: Dice coefficient, \textit{Right}: ASSD score. Held-out domain: Lumbar.}
\end{figure}

\begin{figure}[H]
    \centering
    \subfigure{%
    \includegraphics[height=1.5in, width=2.2in]{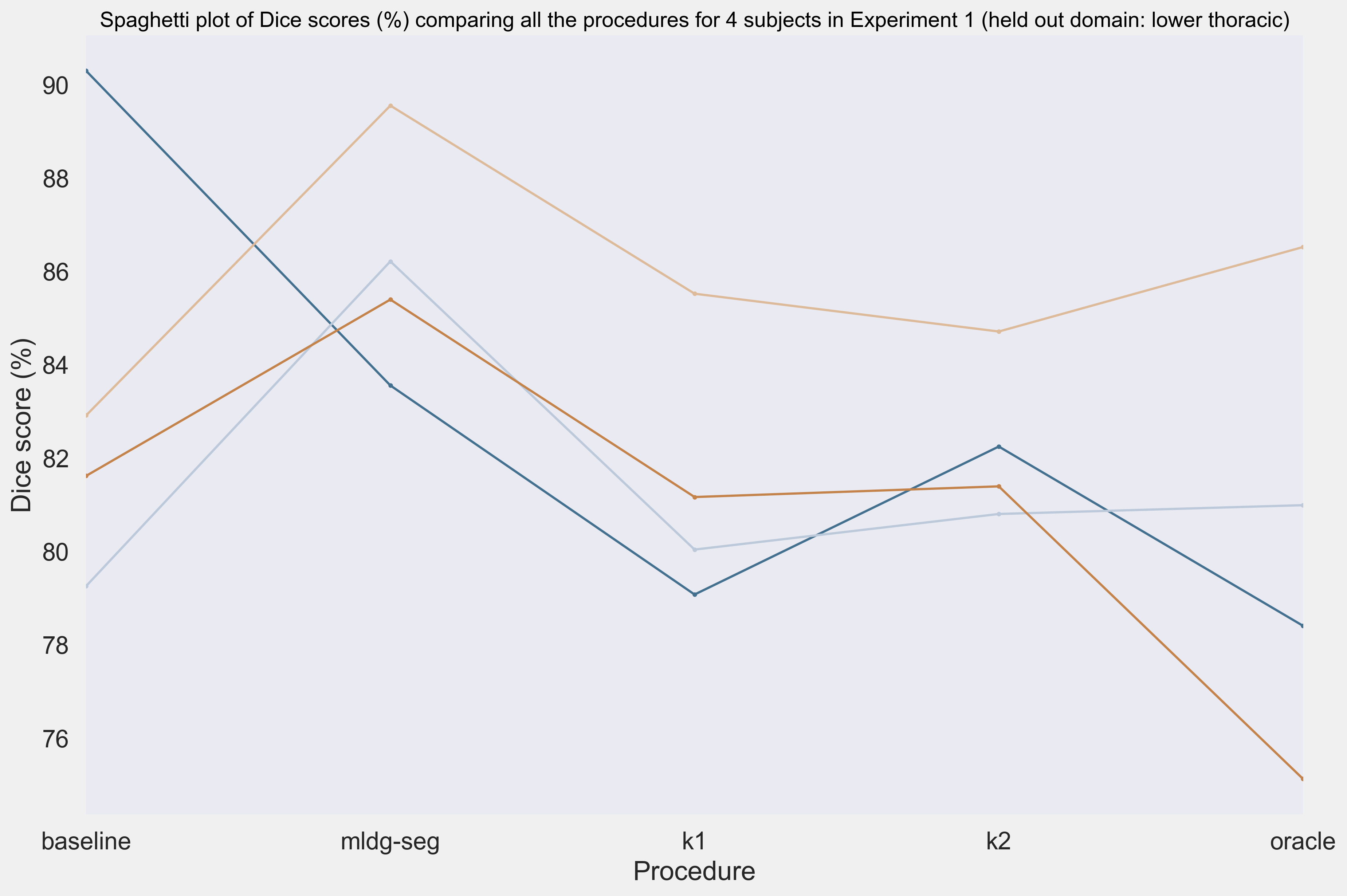}}%
    \qquad
    \subfigure{%
    \includegraphics[height=1.5in, width=2.2in]{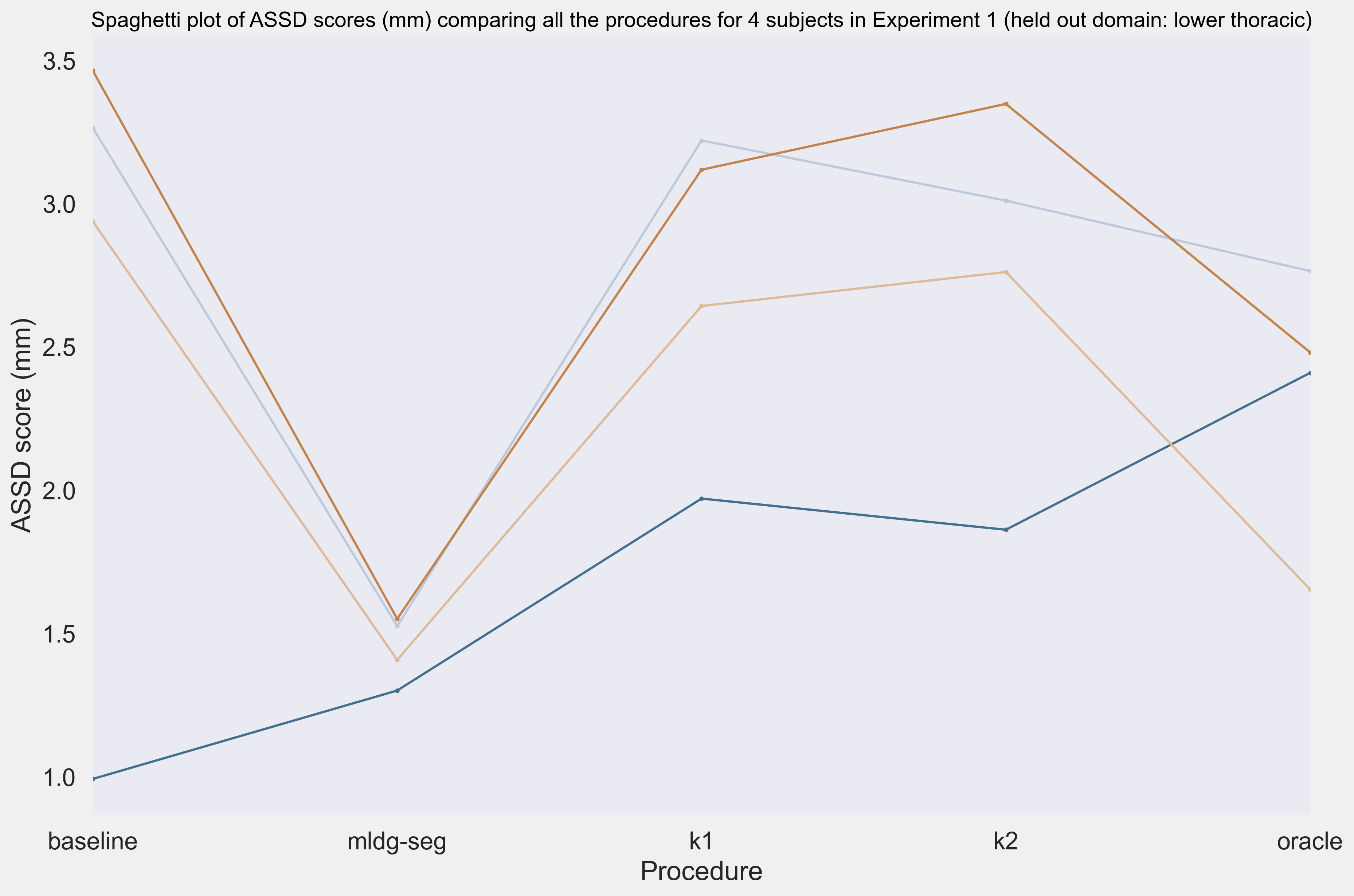}}%
    \caption{\textit{Left}: Dice coefficient, \textit{Right}: ASSD score. Held-out domain: Lower thoracic.}
\end{figure}

\begin{figure}[H]
    \centering
    \subfigure{%
    \includegraphics[height=1.5in, width=2.2in]{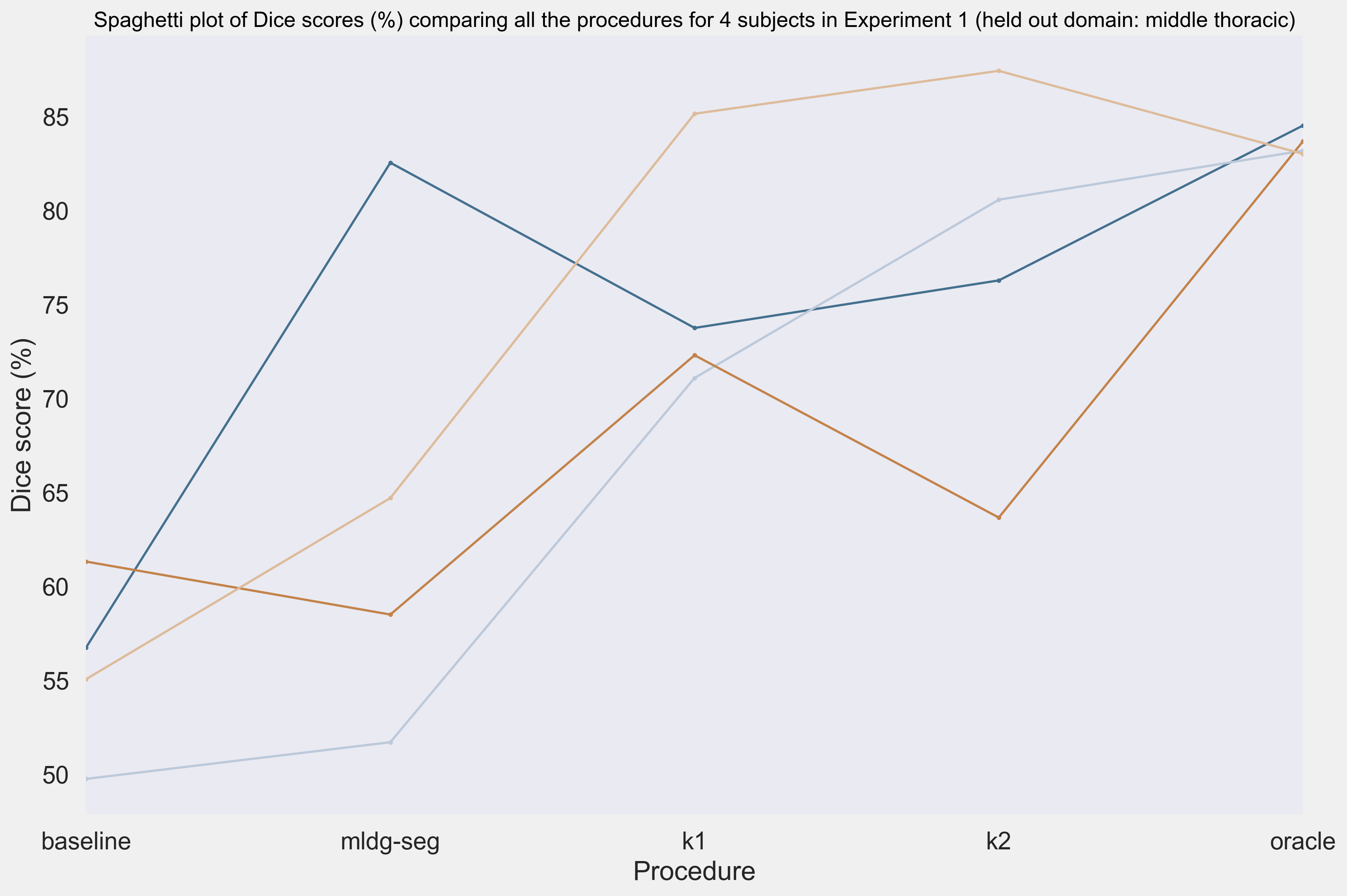}}%
    \qquad
    \subfigure{%
    \includegraphics[height=1.5in, width=2.2in]{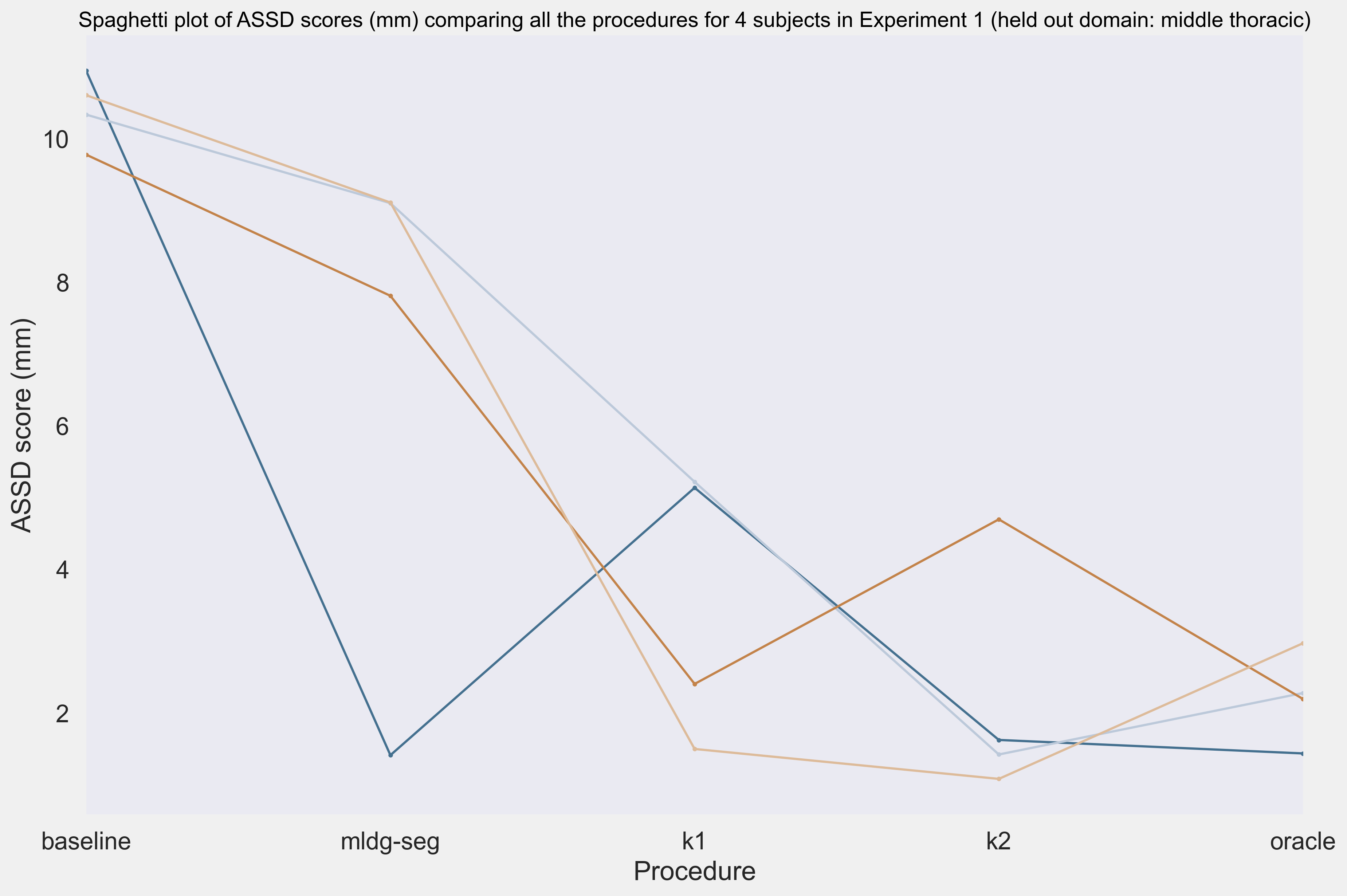}}%
    \caption{\textit{Left}: Dice coefficient, \textit{Right}: ASSD score. Held-out domain: Middle thoracic.}
\end{figure}

\begin{figure}[H]
    \centering
    \subfigure{%
    \includegraphics[height=1.5in, width=2.2in]{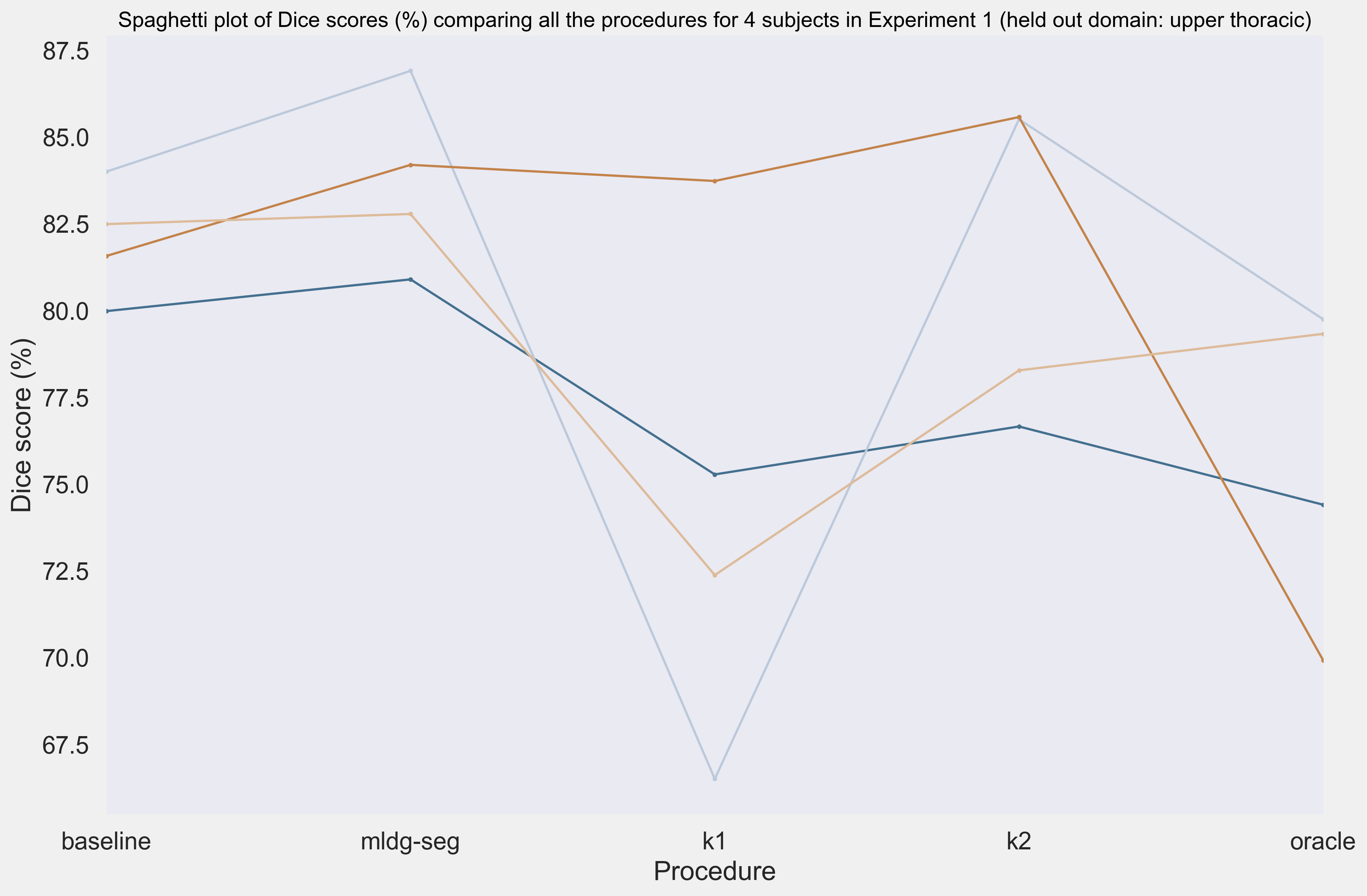}}%
    \qquad
    \subfigure{%
    \includegraphics[height=1.5in, width=2.2in]{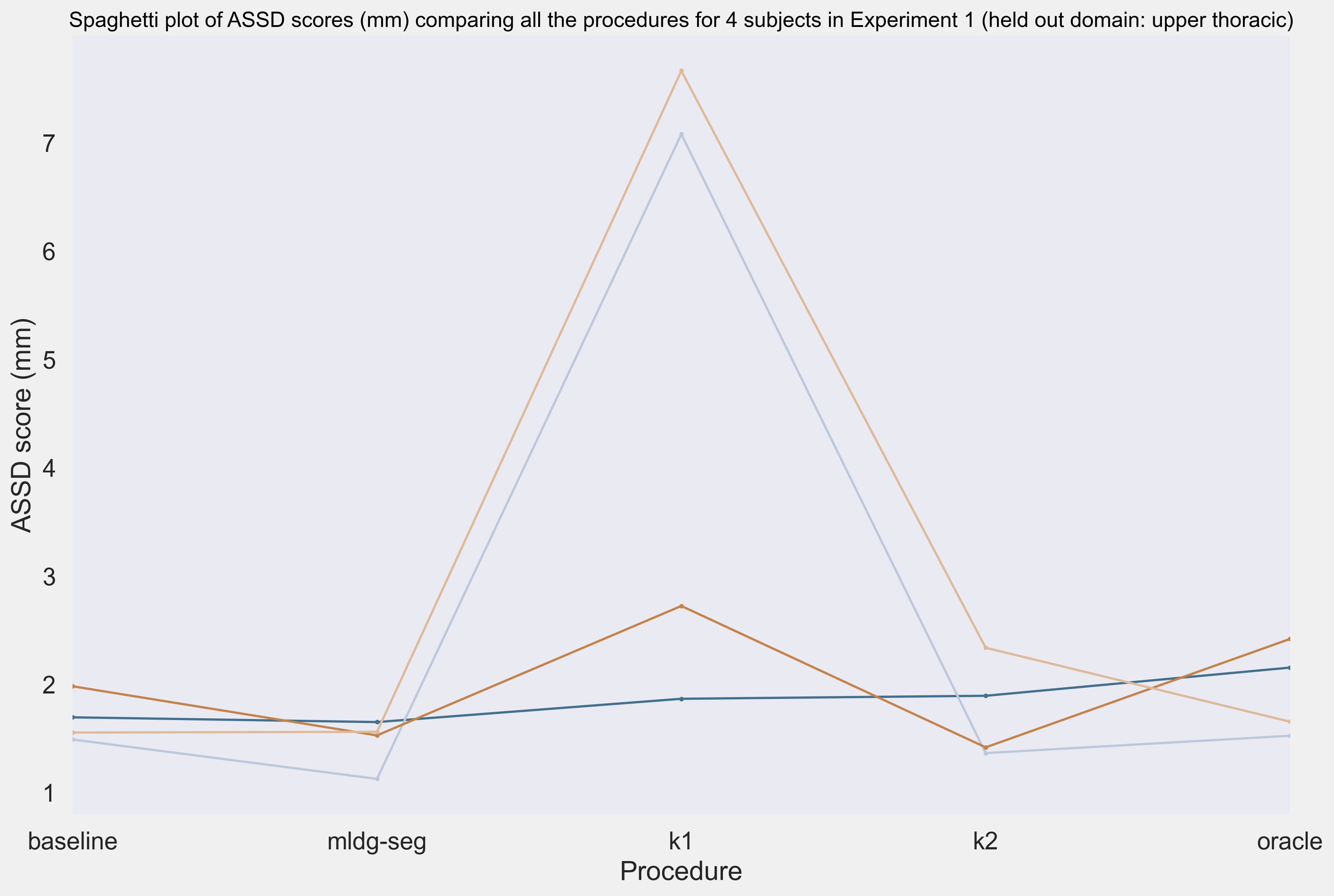}}%
    \caption{\textit{Left}: Dice coefficient, \textit{Right}: ASSD score. Held-out domain: Upper thoracic.}
\end{figure}

\subsection{Experiment 1 (10 subjects) [Refer Table 1 in this supplement in Section \ref{last} above]}
\label{spaghettiExp1_10}

\begin{figure}[H]
    \centering
    \subfigure{%
    \includegraphics[height=1.5in, width=2.2in]{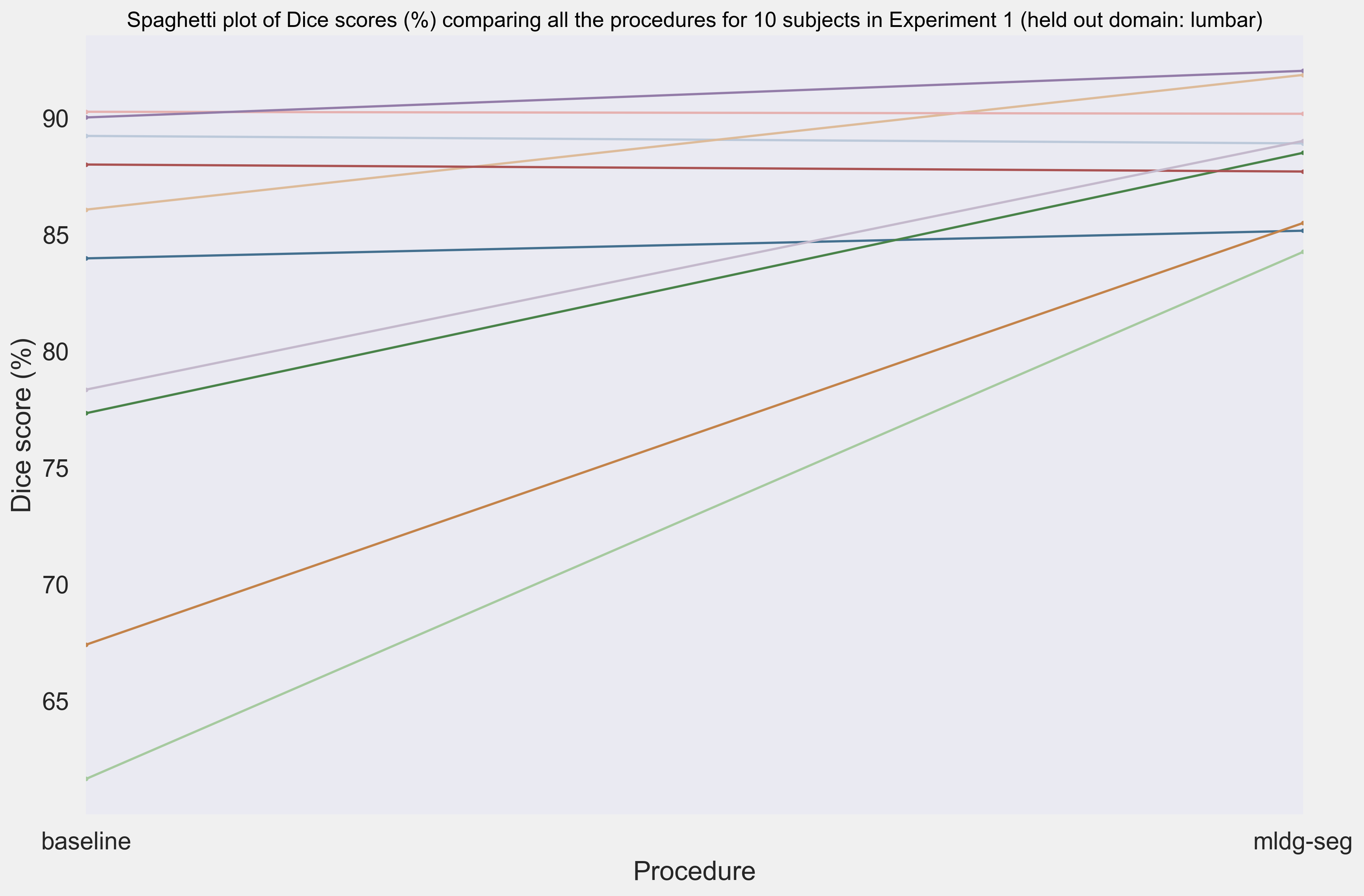}}%
    \qquad
    \subfigure{%
    \includegraphics[height=1.5in, width=2.2in]{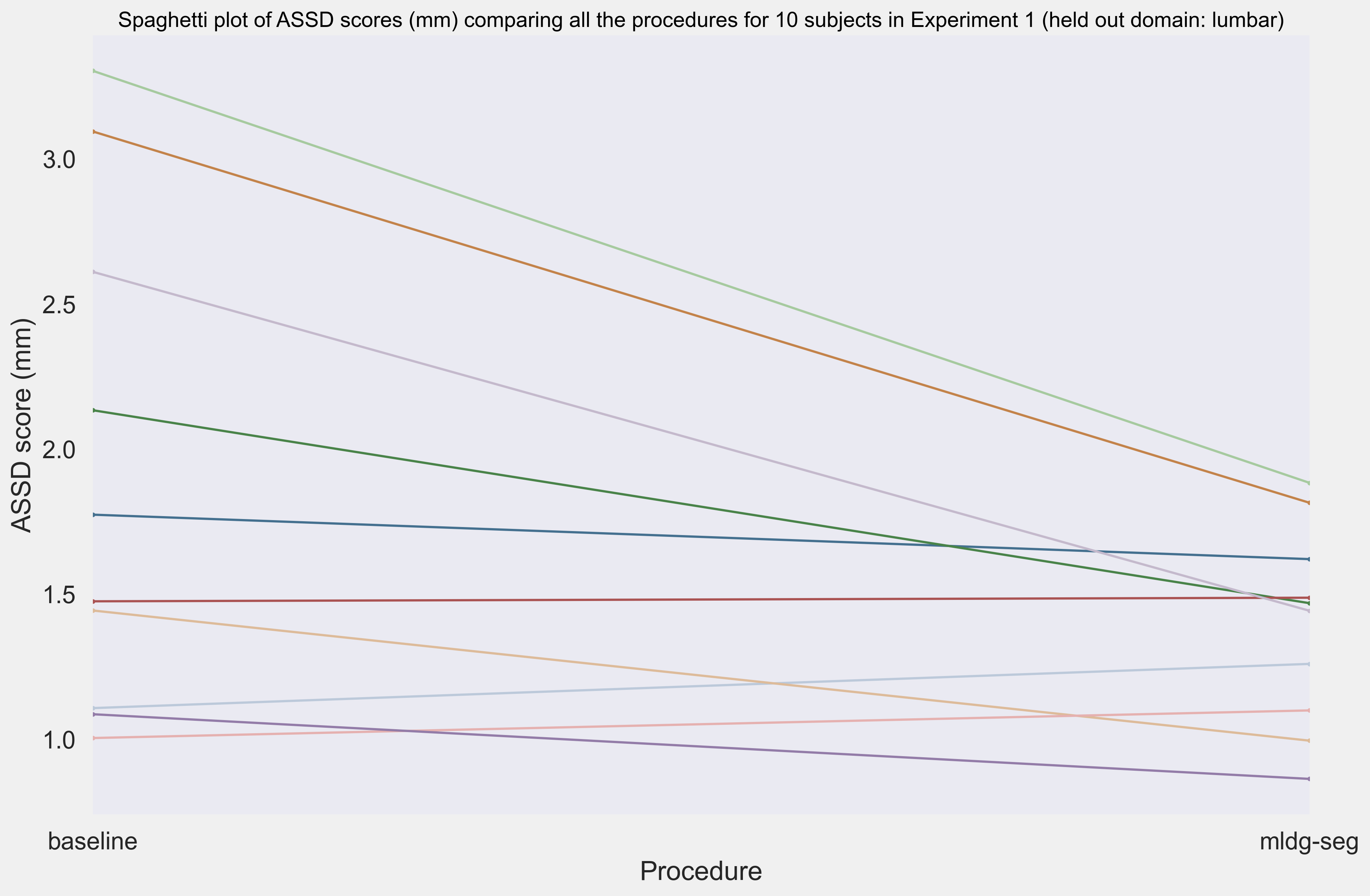}}%
    \caption{\textit{Left}: Dice coefficient, \textit{Right}: ASSD score. Held-out domain: Lumbar.}
\end{figure}

\begin{figure}[H]
    \centering
    \subfigure{%
    \includegraphics[height=1.5in, width=2.2in]{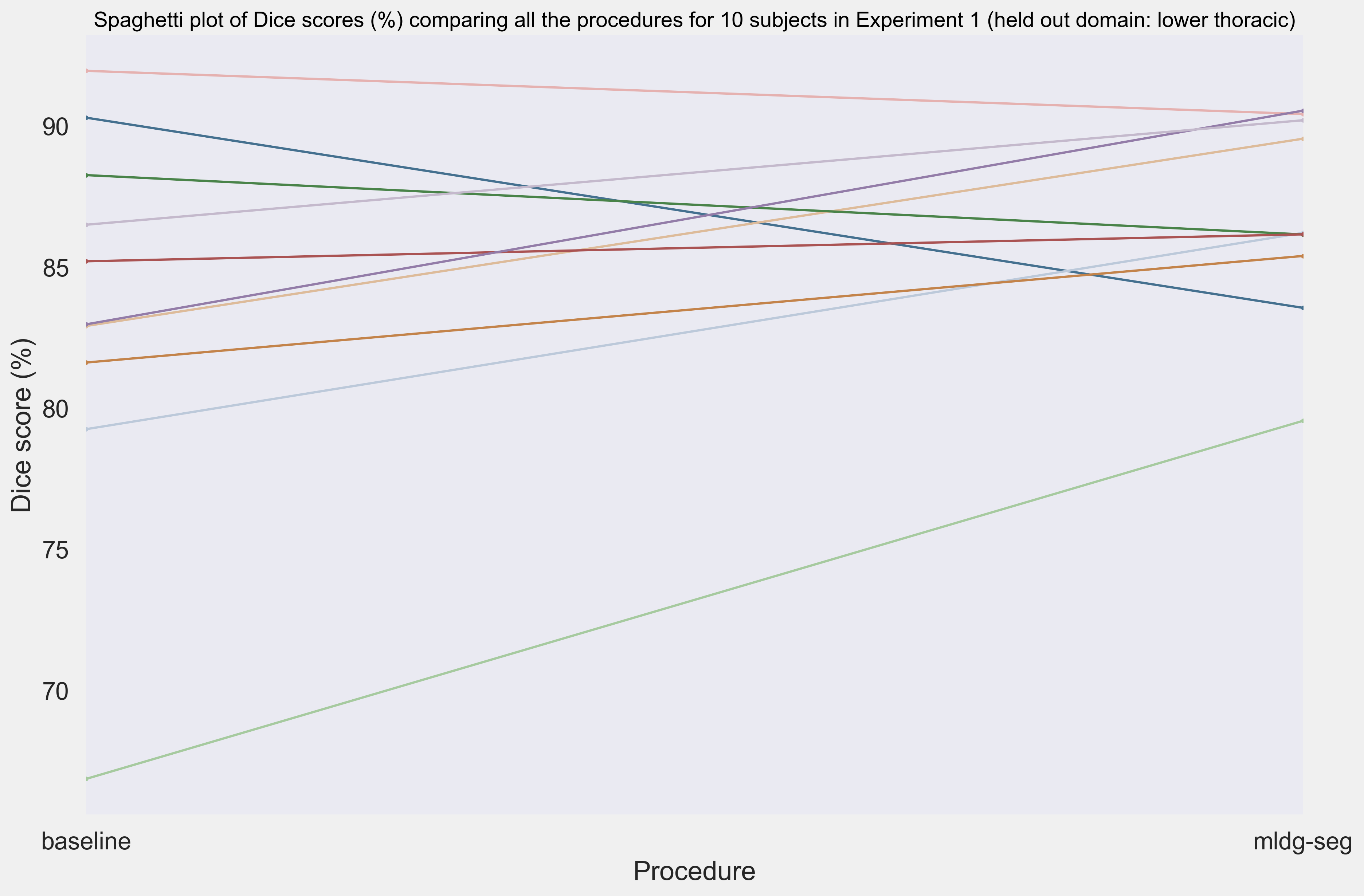}}%
    \qquad
    \subfigure{%
    \includegraphics[height=1.5in, width=2.2in]{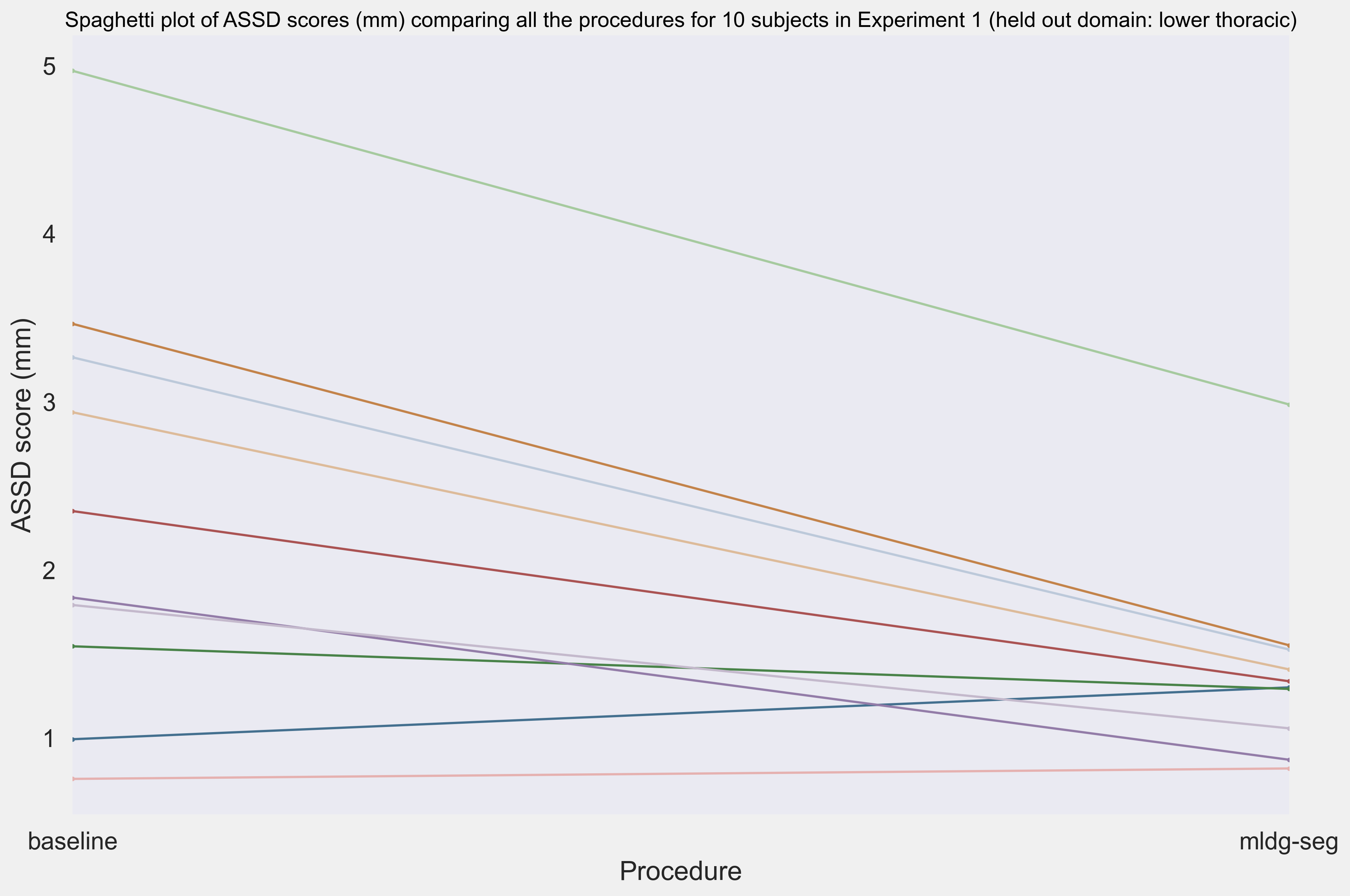}}%
    \caption{\textit{Left}: Dice coefficient, \textit{Right}: ASSD score. Held-out domain: Lower thoracic.}
\end{figure}

\begin{figure}[H]
    \centering
    \subfigure{%
    \includegraphics[height=1.5in, width=2.2in]{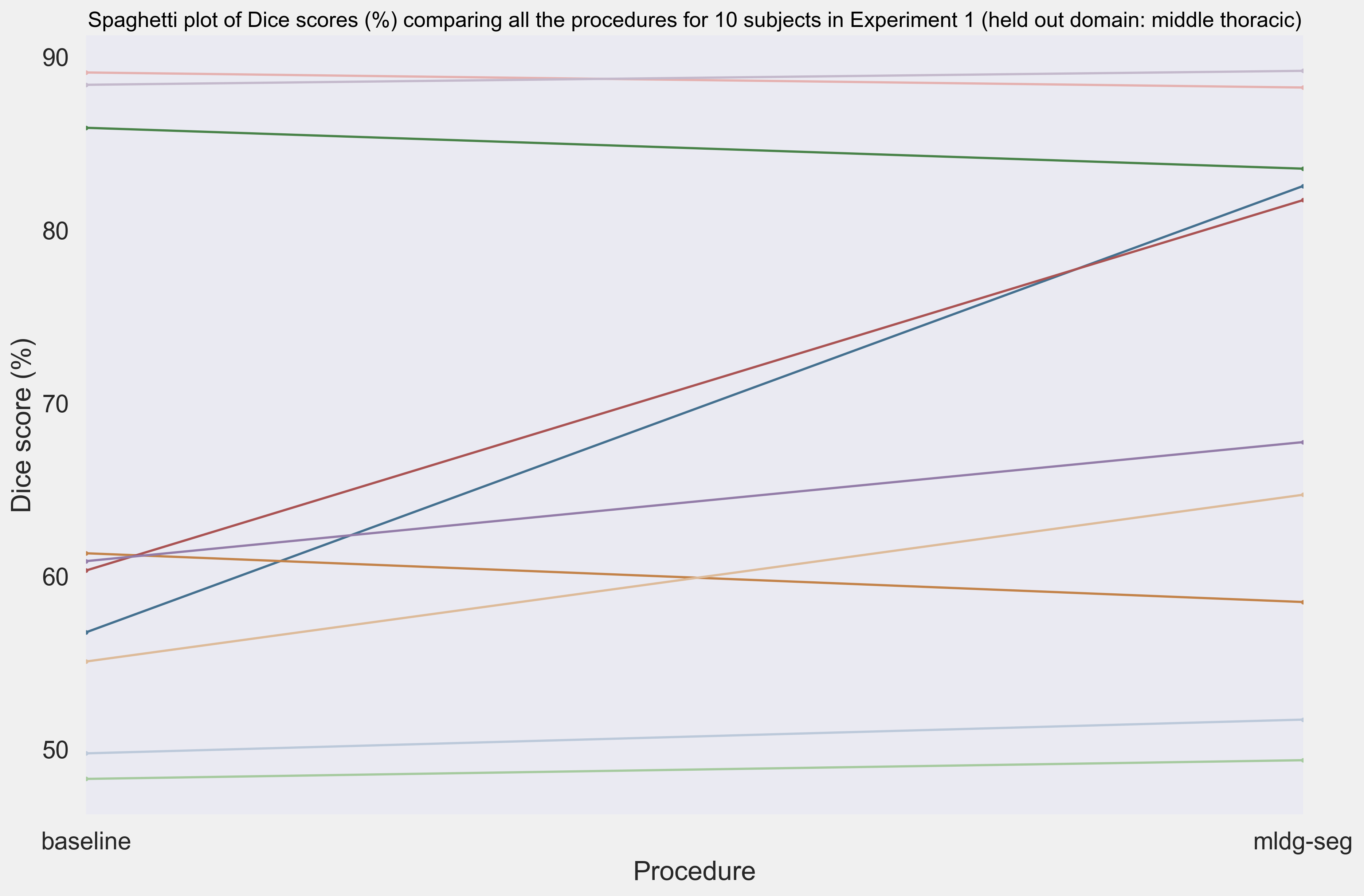}}%
    \qquad
    \subfigure{%
    \includegraphics[height=1.5in, width=2.2in]{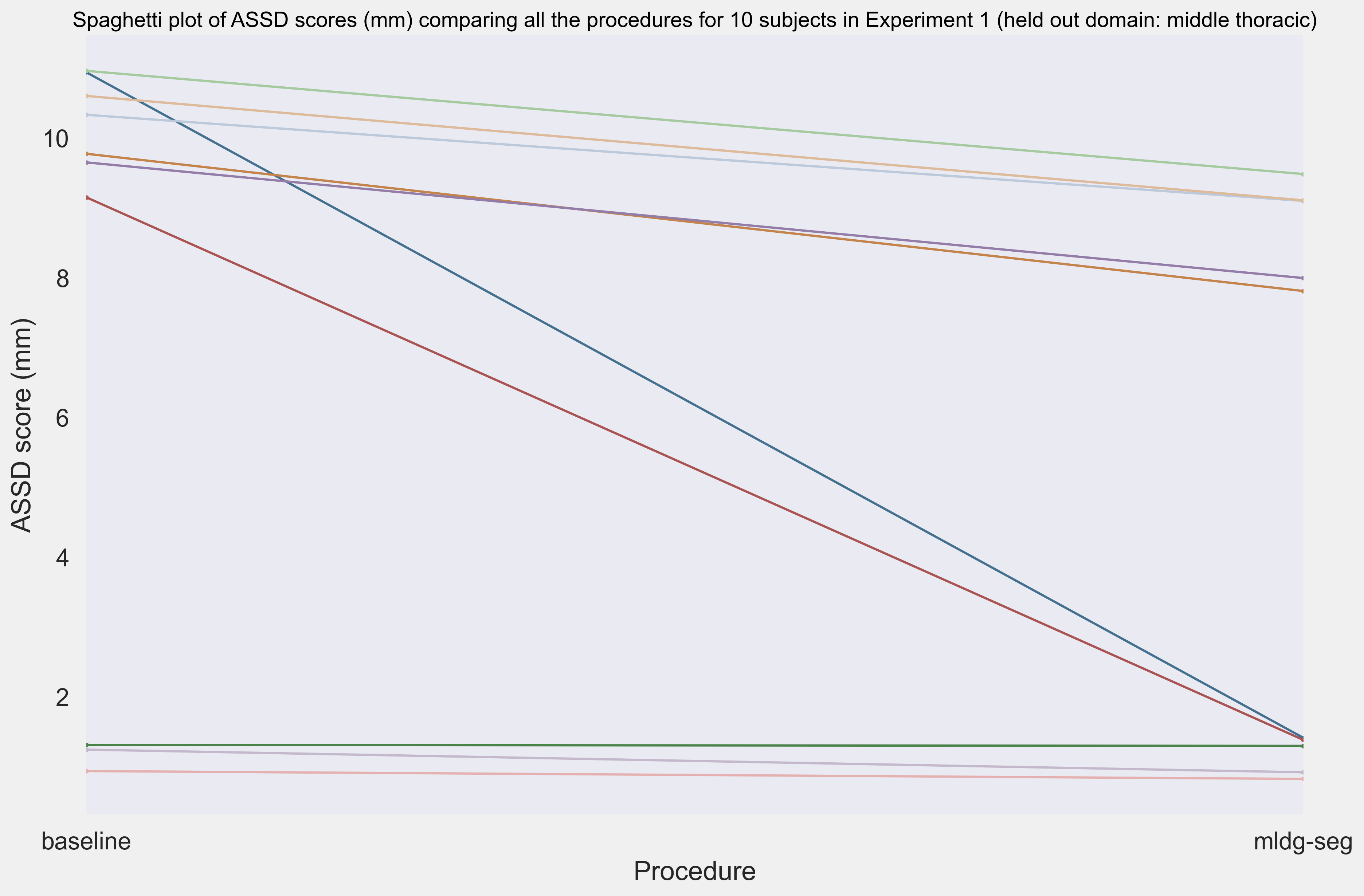}}%
    \caption{\textit{Left}: Dice coefficient, \textit{Right}: ASSD score. Held-out domain: Middle thoracic.}
\end{figure}

\begin{figure}[H]
    \centering
    \subfigure{%
    \includegraphics[height=1.5in, width=2.2in]{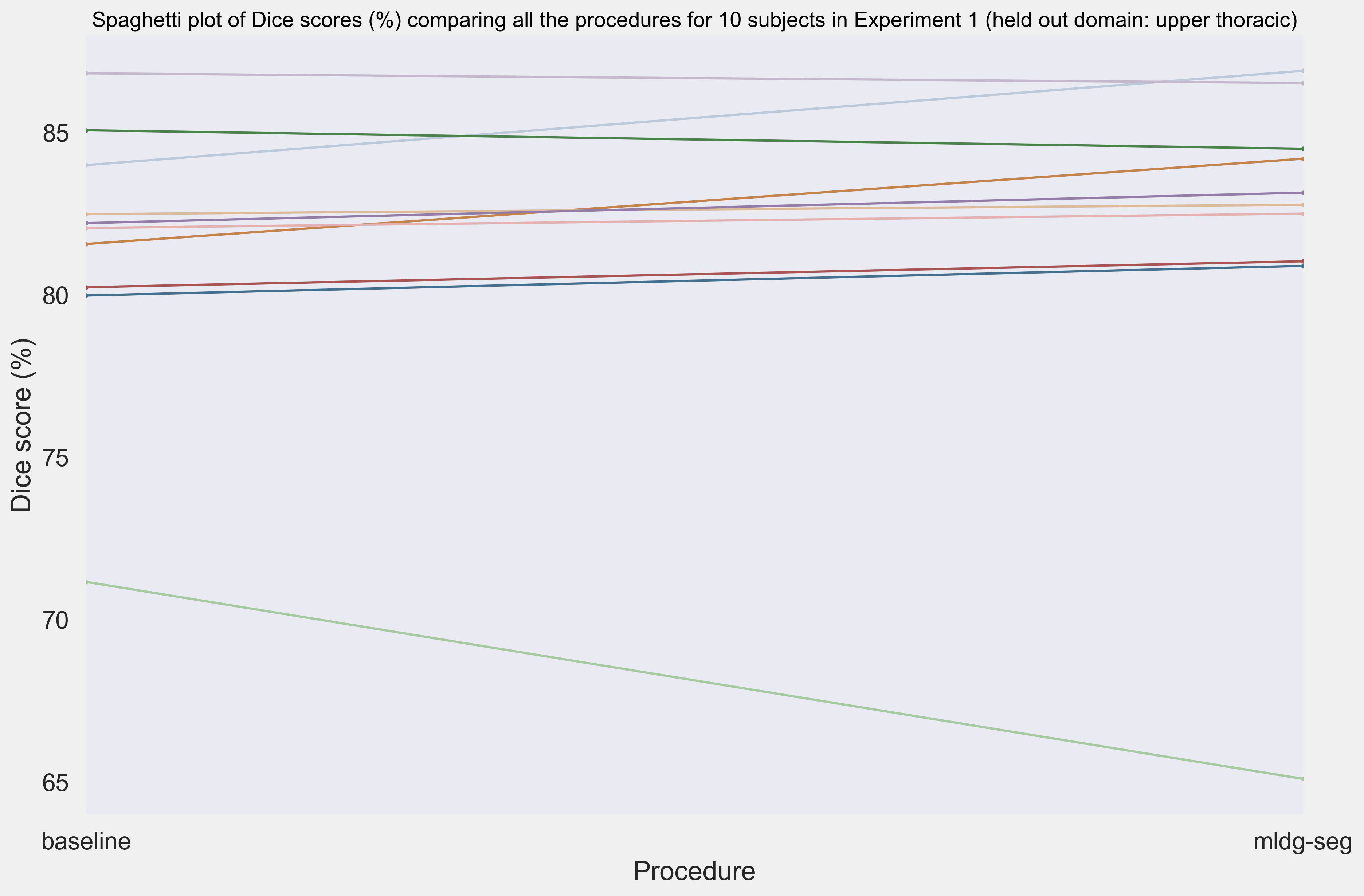}}%
    \qquad
    \subfigure{%
    \includegraphics[height=1.5in, width=2.2in]{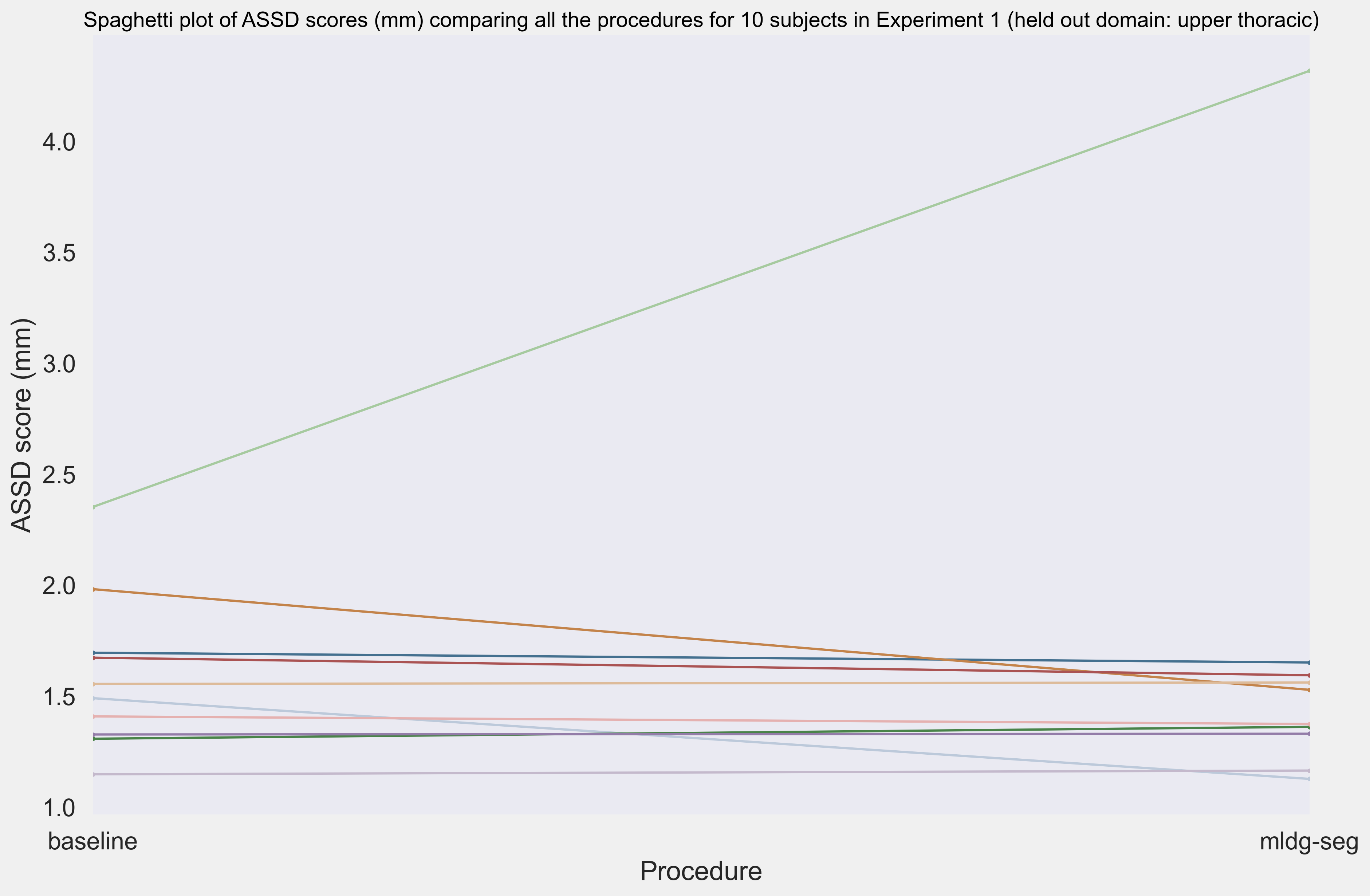}}%
    \caption{\textit{Left}: Dice coefficient, \textit{Right}: ASSD score. Held-out domain: Upper thoracic.}
\end{figure}

\subsection{Experiment 2 [Refer Table 2 in the main paper]}
\label{spaghettiExp2}

\begin{figure}[H]
    \centering
    \subfigure{%
    \includegraphics[height=1.5in, width=2.2in]{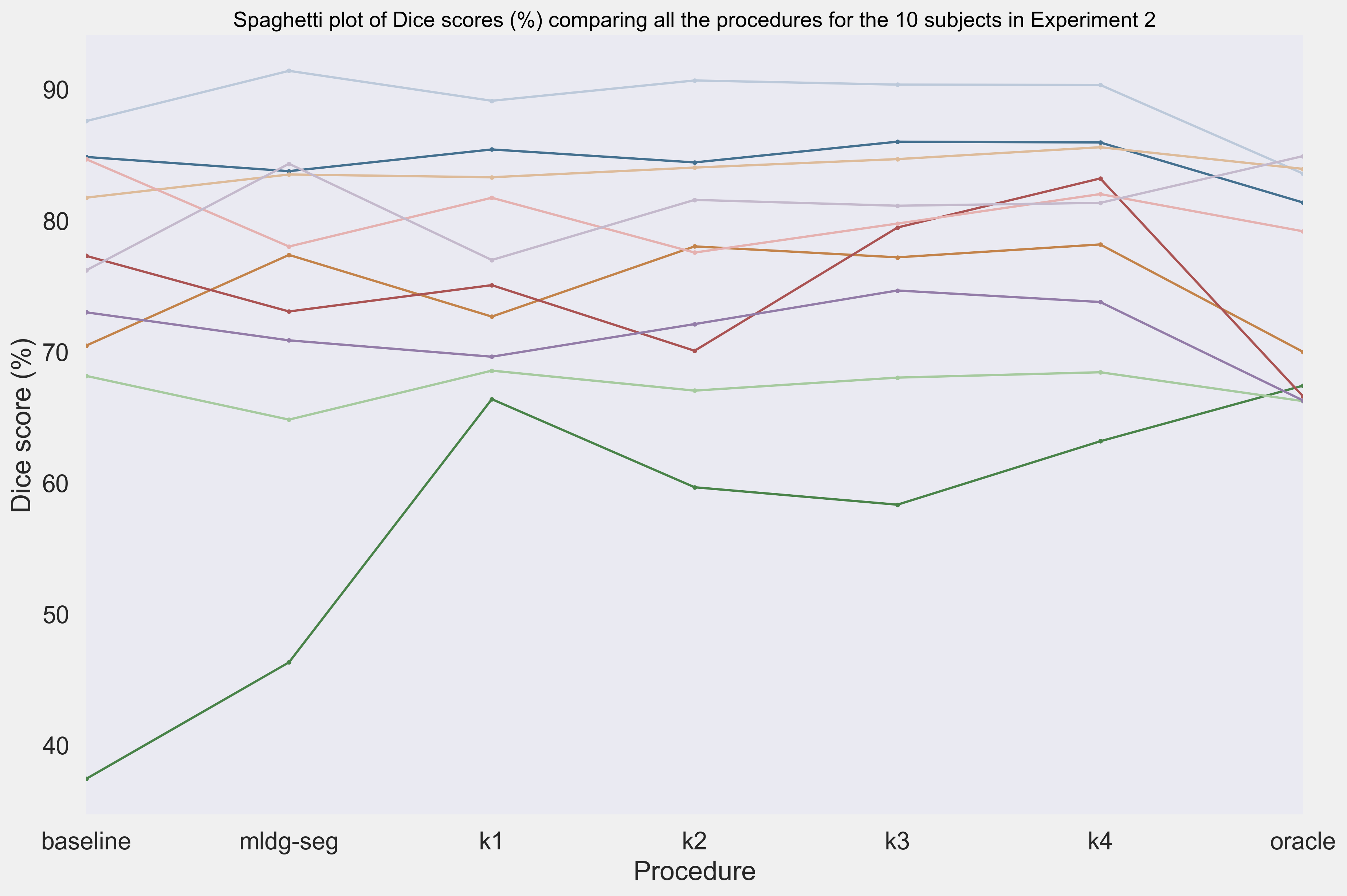}}%
    \qquad
    \subfigure{%
    \includegraphics[height=1.5in, width=2.2in]{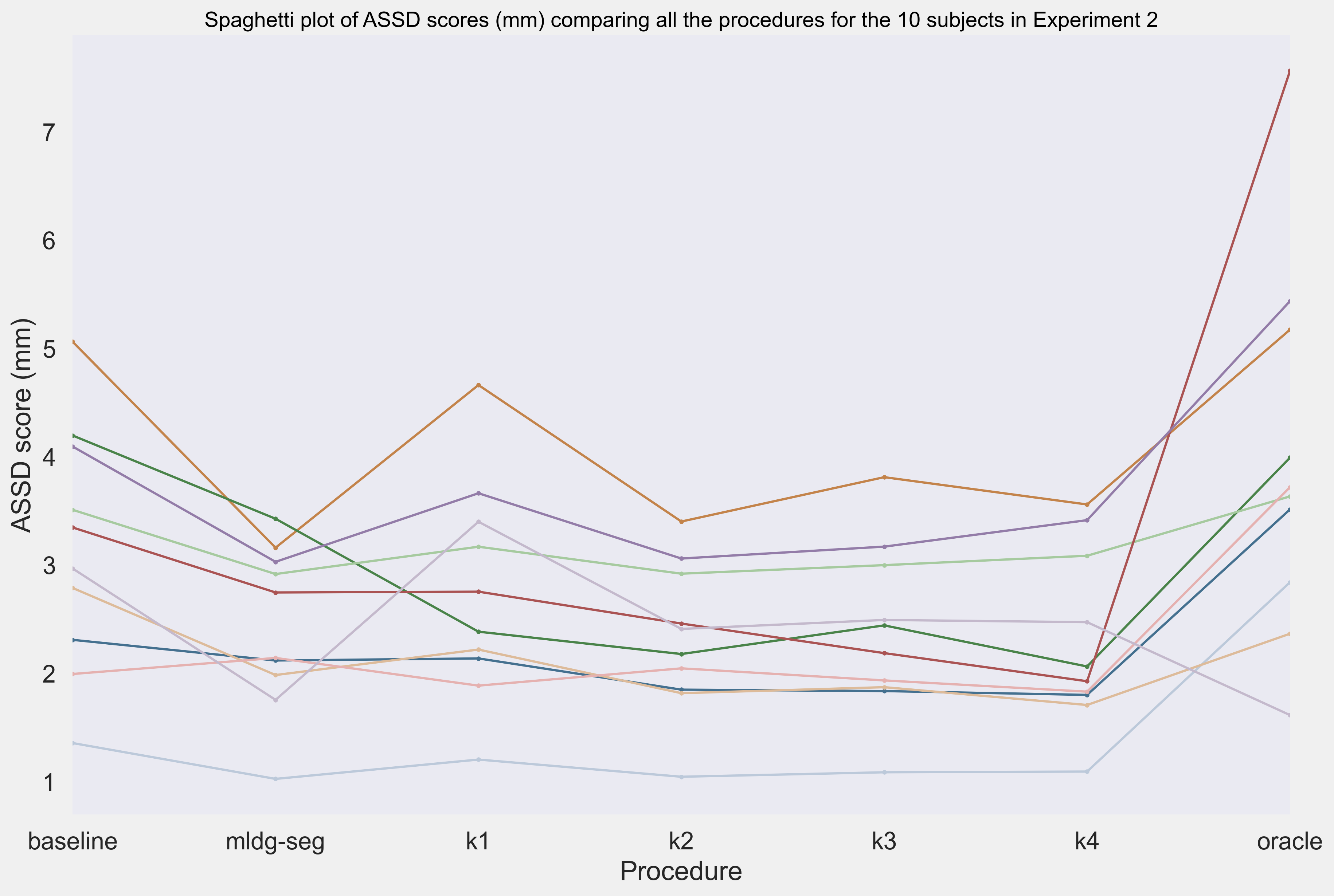}}%
    \caption{\textit{Left}: Dice coefficient, \textit{Right}: ASSD score. Experiment 2.}
\end{figure}

\end{document}